\documentclass[sigconf]{acmart}
\usepackage{bm}
\usepackage{hyperref}
\DeclareMathOperator*{\argmin}{argmin}
\AtBeginDocument{%
  \providecommand\BibTeX{{%
    \normalfont B\kern-0.5em{\scshape i\kern-0.25em b}\kern-0.8em\TeX}}}

\copyrightyear{2020}
\acmYear{2020}
\setcopyright{acmcopyright}\acmConference[MM '20]{Proceedings of the 28th ACM International Conference on Multimedia}{October 12--16, 2020}{Seattle, WA, USA}
\acmBooktitle{Proceedings of the 28th ACM International Conference on Multimedia (MM '20), October 12--16, 2020, Seattle, WA, USA}
\acmPrice{15.00}
\acmDOI{10.1145/3394171.3413819}
\acmISBN{978-1-4503-7988-5/20/10}
\begin{document}
\fancyhead{}
\title{TextRay: Contour-based Geometric Modeling for Arbitrary-shaped Scene Text Detection}

%
% The "author" command and its associated commands are used to define
% the authors and their affiliations.
% Of note is the shared affiliation of the first two authors, and the
% "authornote" and "authornotemark" commands
% used to denote shared contribution to the research.
\author{Fangfang Wang}
\affiliation{\institution{College of Computer Science, Zhejiang University}}
\email{fangfangliana@zju.edu.cn}

\author{Yifeng Chen}
\affiliation{\institution{College of Computer Science, Zhejiang University}}
\email{yifengchen@zju.edu.cn}

\author{Fei Wu}
\affiliation{\institution{College of Computer Science, Zhejiang University}}
\email{wufei@zju.edu.cn}

\author{Xi Li}
\authornote{Corresponding author.}
\affiliation{\institution{College of Computer Science, Zhejiang University}}
\email{xilizju@zju.edu.cn}

%
% By default, the full list of authors will be used in the page
% headers. Often, this list is too long, and will overlap
% other information printed in the page headers. This command allows
% the author to define a more concise list
% of authors' names for this purpose.
\renewcommand{\shortauthors}{Trovato and Tobin, et al.}

%
% The abstract is a short summary of the work to be presented in the
% article.
\begin{abstract}
  Arbitrary-shaped text detection is a challenging task due to the complex geometric layouts of texts such as large aspect ratios, various scales, random rotations and curve shapes.
  Most state-of-the-art methods solve this problem from bottom-up perspectives, seeking to model a text instance of complex geometric layouts with simple local units (e.g., local boxes or pixels) and generate detections with heuristic post-processings.
  In this work, we propose an arbitrary-shaped text detection method, namely TextRay, which conducts top-down contour-based geometric modeling and geometric parameter learning within a single-shot anchor-free framework.
  The geometric modeling is carried out under polar system with a bidirectional mapping scheme between shape space and parameter space, encoding complex geometric layouts into unified representations.
  For effective learning of the representations, we design a central-weighted training strategy and a content loss which builds propagation paths between geometric encodings and visual content.
  TextRay outputs simple polygon detections at one pass with only one NMS post-processing.
  %It bridges the gap between arbitrarily distributed appearance and unified contour-based representation for scene text detection.
  Experiments on several benchmark datasets demonstrate the effectiveness of the proposed approach.
  The code is available at \color{ACMDarkBlue}{\href{https://github.com/LianaWang/TextRay}{https://github.com/LianaWang/TextRay}}.
\end{abstract}




%%% A "teaser" image appears between the author and affiliation
%%% information and the body of the document, and typically spans the
%%% page.

\maketitle

\section{Introduction}
In recent years, scene text detection has attracted a surge of research interest in computer vision community for its wide-range applications such as autonomous driving, blind navigation and scene parsing.
The task of scene text detection aims to localize text areas in natural images.
Despite the significant achievements in object detection, accurately detecting scene texts remains challenging due to their unique traits in geometric layouts: large aspect ratios, various scales, random rotations, and curve shapes.

\begin{figure}[t]
\begin{center}
\includegraphics[width=1.0\linewidth]{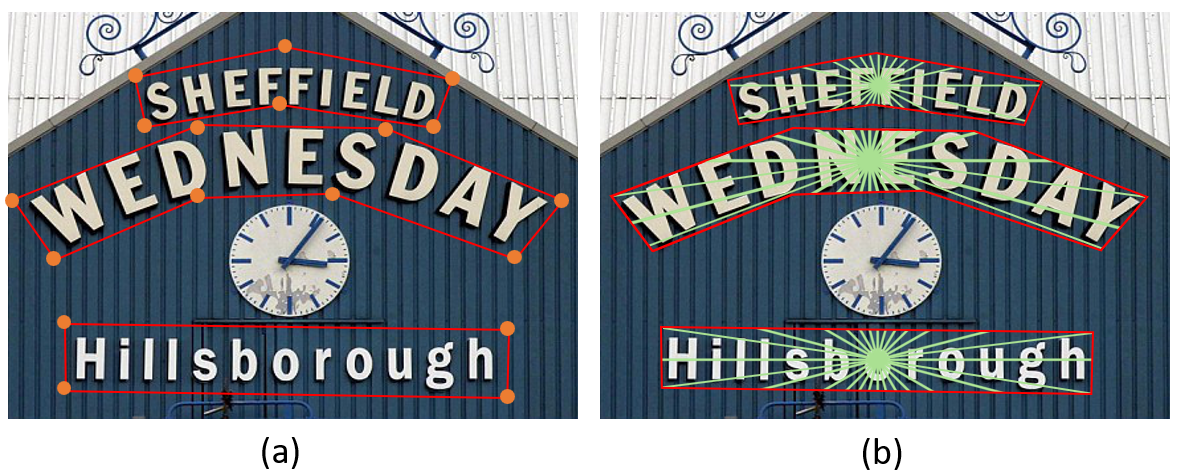}
\end{center}
\vspace{-1.5em}
   \caption{Demonstration of different global representations. (a) shows the global representation under Cartesian system: the three instances are represented with variable-length points (e.g., 6, 8 and 4 orange dots) with very subjective distribution. (b) shows the global representation under polar system: all instances are represented by a set of radiuses of same number and angular distribution (green rays).}
\label{fig:demonstration}
\vspace{-1.5em}
\end{figure}

Different from generic object detection which requires bounding-box outputs, scene text detection demands explicit contours of text instances.
With this motivation, a lot of efforts have been paid in seeking effective contour modeling for arbitrary-shaped text representation.
A typical way adopted in state-of-the-art methods is to model text instances from local perspectives.
These local modeling methods usually decompose a text instance with complex geometric layout into simple control units.
For example, ~\cite{textsnake, feng2019textdragon} represent text instances with sequential discs and local boxes, respectively.
Segmentation-based methods~\cite{wang2019efficient,wang2019shape}, which model arbitrary-shaped text instances at pixel level and represent the contours with the edge of masks, can be seen as an extreme case of local modeling method.

However, flexible as local modeling methods are, they tend to be less resistant to noises and dependant on heuristic post-processings in contour generation due to the homogenous texture within text regions.
More importantly, they commonly pay less attention to the global geometric distribution which reflects the integrated layouts of the contours.
In current benchmarks, the global layouts of arbitrary-shaped text contours are represented with adaptive or fixed number of vertices under Cartesian system, as shown in Figure~\ref{fig:demonstration}(a).
But the vertices are usually not uniformly distributed and independent to each other, neglecting the constraints of global geometric layout.
Besides, the nonunified representations of adaptive number of vertices are unsuitable for parameter learning within unified convolutional neural networks(CNNs).
Based on these observations, we find out that a proper global modeling method which encodes contour-level geometric information into standardized representations is needed to enable unified parameter learning.

So in this paper, we propose an arbitrary-shaped scene text detection method, namely TextRay, which conducts contour-based global geometric modeling and end-to-end parameter learning within a unified framework.
Inspired by recent instance segmentation work ESE-Seg~\cite{xu2019explicit}, we model the contours under polar system.
As illustrated in Figure~\ref{fig:demonstration}(b), a polar system is constructed at each text center, then by emitting fixed number of rays from the text centers according to the same angular distribution and intersecting the contours, we obtain unified representations for different instances in the form of sets of radiuses.
This unified representation can be directly regressed with common CNNs, but it does not fully exploit the intrinsic geometric patterns, for example long patterns and symmetry properties.
To dig up into the geometric patterns lying in the distribution of radiuses, we abstract a higher level geometric encoding by fitting the function curve of radiuses with respect to angles.
In this way, the arbitrary-shaped contours are mapped from complex shape space to unified parameter space, and the unified fitting parameters (i.e., geometric encodings) intrinsically encodes the geometric pattern information.
In this work, we represent arbitrary text contours with unified geometric encodings.
To make the abstract geometric encodings more learnable, we propose a bidirectionally differentiable mapping scheme between shape space and parameter space to measure the distance of geometric encodings according to visual content.
% Given a unique geometric encoding, a unique contour can be recovered by simple and fast calculations.

In principle, the TextRay is a single-shot, anchor-free, and fully-convolutional framework that performs joint optimization between text/non-text classification and geometric encodings regression.
It adopts an end-to-end central-weighted training strategy to effectively deal with long instances which are prevalent in scene texts.
TextRay densely predicts geometric encodings and outputs simple polygon detections at one pass with only one NMS post-processing step.
The main contributions of this paper are as follows:
\begin{itemize}
\item we convert the complex text contour discovery problem into a simple parameter learning problem through global geometric modeling and propose a bidirectionally differentiable mapping scheme between shape space and parameter space for effective geometric parameter learning;
\end{itemize}
\begin{itemize}
\item we present a single-shot anchor-free framework with central-weighted training strategy for solving the arbitrary-shaped scene texts detection problem;
\end{itemize}
\begin{itemize}
\item as a light-weighted one-stage regression-based detector, the TextRay outperforms many segmentation-based or multi-stage detectors and achieves competitive results.
\end{itemize}

\section{Related Work}
Over the past few years, the mainstream of scene text detection approaches has altered from traditional methods to deep learning methods.
Many CNN-based methods are inspired by general object detection and segmentation frameworks like Faster-rcnn~\cite{faster-rcnn}, SSD~\cite{SSD}, Mask-rcnn~\cite{maskrcnn} and FCN~\cite{fcn}.
These methods can mainly be divided into two groups according to their modeling perspectives for text instances: bottom-up local modeling methods and top-down global modeling methods.

\textit{\textbf{Local modeling methods.}}
Local modeling methods usually decompose scene text instances into pixels or fragments.
For example, segmentation-based methods~\cite{multi-channel,multi-orient,masktextspotter} usually explore text region at pixel-level.
Segmenting text instances without cluttering or missing of parts is not easy because scene texts often appear in parallel and the texture of texts is highly homogeneous.
To tackle this problem, affinity information~\cite{textfield,baek2019character,deng2018pixellink,liu2018learning} or multi-level segmentation~\cite{textsnake,wang2019efficient,wang2019shape} are utilized to guide text saliency map generation.
Another typical local modeling method is the combined method that borrows techniques from both segmentation and regression methodology.
Most of these methods seek pixel-level information discovery and supportive cues from object knowledge.
~\cite{cornerregion} combines region segmentation with corner detection.
~\cite{zhang2019look,feng2019textdragon,wang2019single} segment text centerlines to coarsely locate text instances and regress border offsets and local boxes to determine accurate boundaries.
Most recently, ~\cite{contournet} proposes to model the local texture information in orthogonal directions on text proposals and generate the contours by re-scoring two responses.

\textit{\textbf{Global modeling methods.}}
Compared with local modeling methods, global modeling methods directly model scene text instances that possess complex geometric layouts.
These methods take the regression-based methodology and treat texts as a special type of object, and regress text representations within detection frameworks.
~\cite{wang2019arbitrary} proposes a two-stage detector which extracts text proposals and then sequentially predicts variable-length coordinates with LSTM.
~\cite{abcnet} introduces a concise parametric representation of curved scene text using Bezier curves.
~\cite{itn} encodes global geometric information into affine transformations and extracts geometry-aware features by manipulating the receptive fields.
~\cite{rotationsensitive} introduces rotation-invariant features by rotating convolutional filters for quadrilateral vertex offset regression.
~\cite{geometry_normalize} applies discrete geometry normalization on feature maps to enhance the perception of global geometric distribution.

\begin{figure*}[t]
\begin{center}
\includegraphics[width=0.9\linewidth]{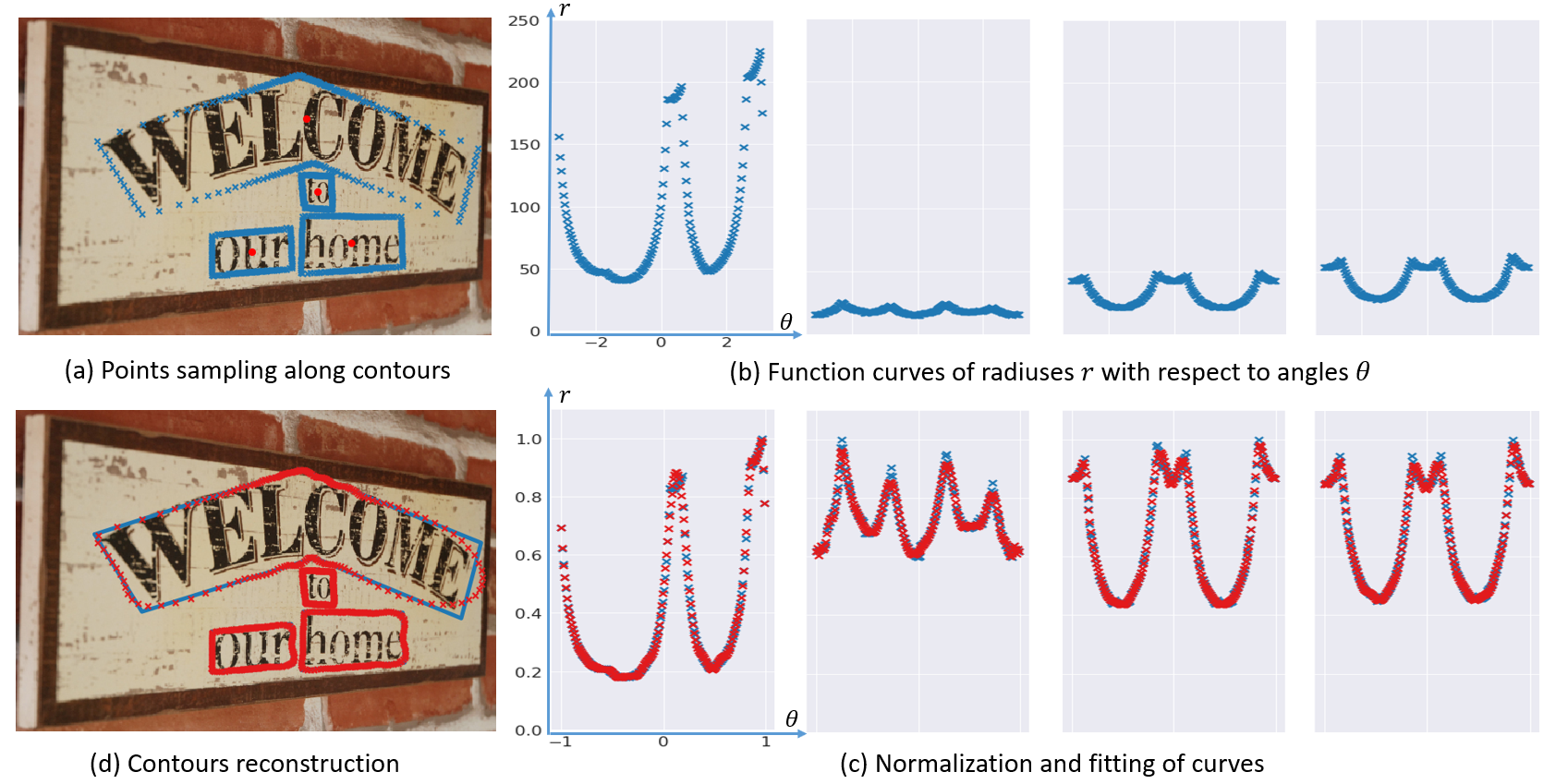}
\end{center}
\vspace{-1.0em}
   \caption{Illustration of the geometric modeling for arbitrary shapes. (a)-(c) shows the geometric encoding process and (d) shows the decoding process. In (a), points (blue crosses) are sampled along text contours by intersecting with $N$ rays emitted from text centers (red dots); (b) visualizes the function curves (blue crosses) of radiuses $\bm{r}$ with respect to angles $\bm{\theta}$; in (c) the function curves are normalized (blue crosses) and fitted (red crosses) with $K$-degree Chebyshev polynomials; (d) displays the ground-truth contours (blue polygon) and the reconstructed contours (red crosses) from the geometric encodings.}
\label{fig:modeling}
\vspace{-1.0em}
\end{figure*}

\textit{\textbf{Discussion.}}
Most local modeling methods are rather flexible and can deal with both quadrilateral and curve text detection.
But this convenience is at the cost of intensive computation and multi-stage processing.
Though regression-based methods achieve high performances on quadrilateral text detection, many of them are unscalable to curve text detection.
So in this paper, we realize a regression-based global modeling method, namely TextRay, for arbitrary-shaped text detection.
In fact, effective contour modeling for irregular objects is also an important research point in general object segmentation tasks~\cite{xie2019polarmask, xu2019explicit,desrep}.
Inspired by~\cite{xu2019explicit}, we model the text contours under polar system with a set of polar radiuses and encode geometric information by fitting the distribution of radiuses with Chebyshev polynomial approximation.
Instead of directly regressing the fitting coefficients in parameter space~\cite{xu2019explicit}, we design a visual content-aware measurement with bidirectionally differentiable space mapping scheme to enable stable high-degree coefficients learning.
Similar to the detection framework~\cite{fcos}, our TextRay is a single-shot anchor-free framework which adopts the FPN~\cite{fpn} and carry out multi-level detection.
Different from~\cite{fcos} which predicts object centerness for false positive suppression, we adopt a central-weighted training strategy to reduce the negative influence of long-shot samples during training stage.

\section{Proposed Method}

The major challenge facing arbitrary-shaped scene text detection is the complex geometric distribution of text instances.% such as various scales, random rotation, large aspect ratio and irregular shape.
In this work, we propose a contour-based geometric modeling method that parameterizes arbitrary-shaped text contours into unified representations (i.e., geometric encodings), and effectively learn the representations within a single-shot anchor-free framework, namely TextRay.

\subsection{Geometric Modeling for Arbitrary Shapes}
\label{sec:geometric}

Texts in natural scenes often appear in arbitrary shapes.
Instead of adopting the intuitive representation with independent vertices in Cartesian coordinates, we explore a unified representation that can reflect the intrinsic constrains of their geometric patterns.
Inspired by recent work~\cite{xu2019explicit}, we seek to represent arbitrary shapes of texts under polar system with unified geometric encodings.

Specifically, individual polar systems are constructed for different text instances.
The pole of the polar system is located at the text center, which is defined as the midpoint of the text centerline.
Points are sampled by intersecting the contour with $N$ rays evenly emitted from the text center at angles $\bm{\theta}=[\theta_1, \theta_2, \dots, \theta_N]$ from $-\pi$ to $\pi$  with same interval $\Delta\theta=\frac{2\pi}{N}$ ($N=360$ in this case), as shown in Figure~\ref{fig:modeling}(a).
The radiuses $\bm{r}=[r_1, r_2, \dots, r_N]$ are defined as the distances between the sampled points and the text center.
Note that when there are multiple intersected points, we keep the farthest one to ensure the integrity of the contour.
The polar coordinates of the sampled points $\{(\theta_i, r_i)\}_{i=1}^N$ forms a function of radiuses $\bm{r}$ with respect to angles $\bm{\theta}$, as shown in Figure~\ref{fig:modeling}(b).
When the angular distribution is fixed, the arbitrary shapes can be represented by $\bm{r}$.
However, elements in $\bm{r}$ are independent and this direct representation neglects the underlying geometric patterns.
To dig up into the geometric patterns and intrinsic correlations of radiuses, we map the contour from shape space to parameter space by using the Chebyshev approximation to fit the function curve.
The fitting function is defined as:
\begin{equation}\label{eq:chebyapprox}
\begin{aligned}
f_K(\bm{\theta}, \bm{c}) = \sum_{k=0}^{K}{c_k*T_k(\frac{\bm{\theta}}{\pi})},
\end{aligned}
\end{equation}
where $T_k$ are the Chebyshev polynomials of the first kind defined by the recurrence relation
$T_0(x)=1, T_1(x) = x, T_{k}(x) = 2xT_{k-1}(x) - T_{k-2}(x)$, $\pi$ is the normalization factor, and the $K$-degree fitting coefficients $\bm{c}=[c_0, c_1, \dots, c_K]$ is defined as the \textit{shape vector} of the instance.
Instead of fitting the original $\bm{r}$ of random scales, we choose to fit the normalized radiuses $\frac{\bm{r}}{s}$ ($s=max(\bm{r})$ is the scale factor for normalization) for balanced coefficients of instances in different scales.
Thus $\bm{c}$ is solved by the least-squares method:
\begin{equation}\label{eq:fitting}
\begin{aligned}
\bm{c} = \argmin_{\bm{c}'}\sum_{i=1}^N
\left(
f_K(\theta_i,\bm{c}') - \frac{r_i}{s}
\right)^2.
\end{aligned}
\end{equation}
The fitted function curve is shown in Figure~\ref{fig:modeling}(c).
Note that this fitting process is with error, which decreases with higher fitting degree $K$.

In this work, we define the scale of an arbitrary-shaped text as the longest sampled radius $s$.
Together with the position $(x, y)$, which are the Cartesian coordinates of the text center, we obtain the unified \textit{geometric encoding} $\bm{ge}=[\bm{c}, s, x, y]$ for the current instance.

Given a unique geometric encoding, we can map it from parameter space back to shape space by reconstructing the contour $\{(x'_i, y'_i)\}_{i=1}^{N}$ with simple calculations:
\begin{equation}\label{eq:resconstruct}
\begin{aligned}
r'_i &= s * f_K(\theta_i, \bm{c}), \\
x'_i &= x + r'_i*cos(\theta_i), \\
y'_i &= y + r'_i*sin(\theta_i). \\
\end{aligned}
\end{equation}
The reconstruction of the contour is depicted in Figure~\ref{fig:modeling}(d).
The whole encoding and decoding process forms the global geometric modeling for arbitrary-shaped text contours.

\subsection{Content Loss for Shape Vector Regression}
\label{sec:content}
Shape vector $\bm{c}$ are parameters that encode the pattern of the arbitrary shape.
Among the geometric properties in the geometric encoding $\bm{ge}=[\bm{c}, s, x, y]$, scale $s$ and position $(x, y)$ are both directly related to the visual content, while $\bm{c}$ is an abstracted algebraic expression of shape pattern.
Directly regressing the shape vector in parameter space will neglect the intra-parameter correlations and fail to make full use of the abundant visual content.
To effectively regress the shape vector, we adopt a differentiable bidirectional mapping scheme with respect to $\bm{c}$ between parameter space and shape space, and incorporate the mapping into a content loss.

\begin{figure}[t]
\begin{center}
\includegraphics[width=1.0\linewidth]{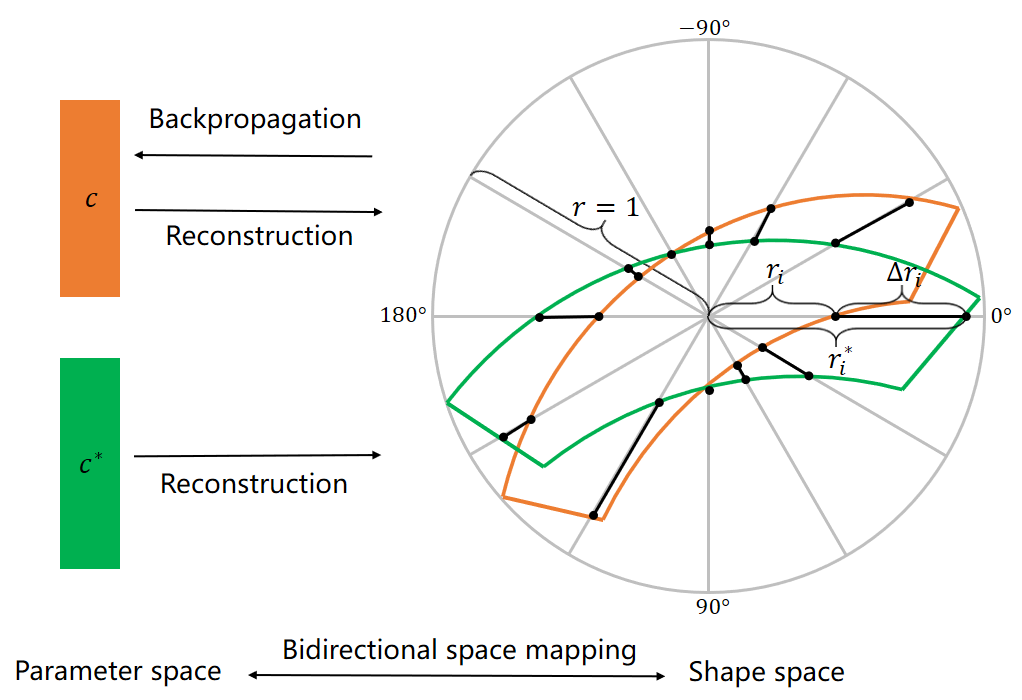}
\end{center}
\vspace{-1.0em}
   \caption{Illustration of the content loss. The orange contour and the green contour are reconstructed from $\bm{c}$ and the ground-truth $\bm{c}^*$ respectively. The distance between the two shape vectors $\bm{c}$ and $\bm{c}^*$ in the parameter space is defined as the average difference between two sets of radiuses in the shape space, that is average length of the black line segments. we only display $12$ radiuses for clarity.}
\label{fig:content}
\vspace{-1.5em}
\end{figure}

As illustrated in Figure~\ref{fig:content}, with the predicted shape vector $\bm{c}$ and ground truth shape vector $\bm{c}^*$, we reconstruct two normalized contours whose $i$-th radius are $r_i = f_K(\theta_i, \bm{c})$ and $r_i^* = f_K(\theta_i, \bm{c}^*)$, respectively.
Scale and position are not considered in this process, so the maximum radiuses and text centers for the two contours are both $1$ and the pole, respectively.
The contours are in the form of $N$ radiuses ($N=360$ in this case).
We measure the distance between shape vectors in the parameter space by means of evaluating the differences between two sets of radiuses in the shape space:
\begin{equation}\label{eq:contentloss}
\begin{aligned}
L_{content}(\bm{c}, \bm{c}^*) = \frac{1}{N} \sum_i^N
smooth_{L_1}
\left(
f_K(\theta_i, \bm{c}) - f_K(\theta_i, \bm{c}^*)
\right),
\end{aligned}
\end{equation}
where smooth-$L_1$ function~\cite{fast-rcnn} is used for regression.

The differentiable content loss builds propagation paths between each radius and each parameter in the shape vector $\bm{c}$.
And thus it is able to effectively model the correlations between the parameterized shape patterns and visual content.

\subsection{Framework of TextRay}
\label{sec:framework}

\begin{figure*}
\begin{center}
\includegraphics[width=0.85\linewidth]{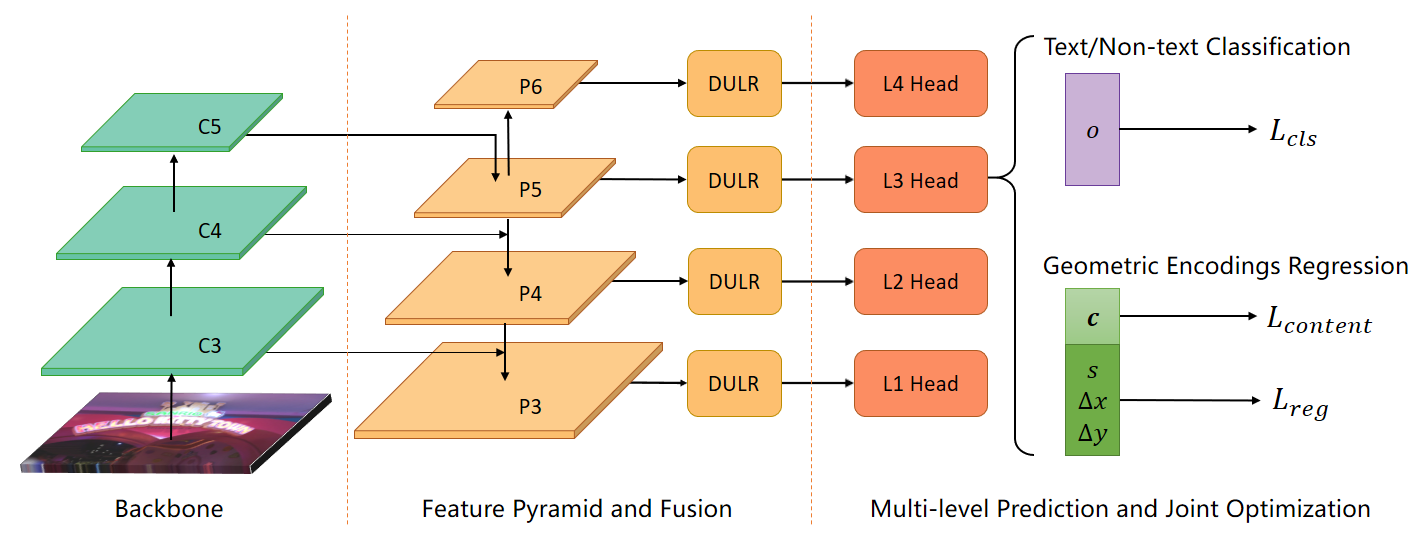}
\end{center}
\vspace{-1.0em}
   \caption{The architecture of TextRay. The TextRay consists of three parts: feature extraction with a backbone network shown on the left; feature pyramid generation and global feature fusion in the middle; and two branch multi-task learning shown on the right where text/non-text classification and geometric encodings regression are jointly optimized on multi-level heads.}
\label{fig:pipeline}
\vspace{-1.5em}
\end{figure*}

As illustrated in Sections ~\ref{sec:geometric} and ~\ref{sec:content}, our global geometric modeling method generates learnable geometric encodings for arbitrary-shaped text instances.
We embed this global geometric modeling into our TextRay framework for parameter learning.

The TextRay is a one-stage fully-convolutional framework that takes an image $\bm{I}$ as input and output simple polygon detections $\bm{D}$.
This concise framework mainly contains three parts: 1) feature extraction; 2) global feature fusion; and 3) two-branch joint optimization on multi-level heads.
The architecture of TextRay is depicted in Figure~\ref{fig:pipeline}.

We utilize ResNet50~\cite{resnet} as our backbone network and adopt FPN~\cite{fpn} to generate multi-level feature maps.
P3, P4 and P5 are derived from C3, C4 and C5 of ResNet50, while P6 is generated by adding a stride two sub-sampling on top of P5.

To ensure sufficient receptive field for long instances, we adopt the SCNN\_DULR (simplified as DULR) module proposed in~\cite{dulr} to generate global features.
In the DULR module, spatial features are propagated in four directions: bottom to top, left to right and their reverse.
Particularly, we replace standard convolution along channel dimension by depth-wise convolution in DULR module to reduce network parameters.
Each used pyramid top is followed by a DULR module and another $3\times3$ convolutional layer.

To tackle the various scales of text instances, we predict targets of different scales at different feature levels similar to FCOS~\cite{fcos}.
The distribution process is on the basis of relative sizes of each text instances with respect to the longer side of the input image.
Different from FCOS~\cite{fcos}, we adopt overlapping size ranges for different level, which means an instance can be simultaneously distributed to more than one level during training.
The overlapping ranges is designed to maintain a higher recall rate.

On top of the feature maps, TextRay incorporates a classification branch and a regression branch.
The classification branch predicts the probability of the current location being text area.
And the regression branch regresses the geometric encodings (i.e., shape, scale and position) of the associated text instance.
The final output $\bm{D}$ is generated by fast decoding according to Equation~\ref{eq:resconstruct} and a Soft-NMS~\cite{soft-nms} step.

\subsection{Central-weighted Training}

To effectively learn the unified geometric encodings and optimize the anchor-free TextRay framework, we design a central-weighted training strategy for scene texts which specializes in dealing with the long structures of text instances.

At training, our anchor-free detector uses points within ground-truth instances on the feature maps to conduct regression tasks.
Points near the ends of a long text instance could be far from the text center, making it hard to accurately sense the comprehensive geometric information.
To guarantee the recall rate and suppress false positives from those long-shot points, we adopt a central-weighted training strategy.

Specifically, points on the feature maps are classified into three categories: positive points that locate in a ground-truth polygon, negative points that are outside of all ground-truth polygons and ignored points that belong to hard instances.
Given a ground-truth text instance whose $\bm{ge}=[\bm{c}, s, x, y]$, the central weight of the $i$-th inside positive point $(x_i, y_i)$ is determined according to its distance from the text center:
\begin{equation}\label{eq:central-weights}
\begin{aligned}
w_i = 1 - \frac{\sqrt{(x_i - x)^2 + (y_i - y)^2}}{s}.
\end{aligned}
\end{equation}
The central weights are applied twice during the training process.
Firstly, the positive point is given a probability $p_i$ of being sampled into a mini-batch according to its central weights:
\begin{equation}\label{eq:sample-mini-batch}
\begin{aligned}
p_i = \frac{w_i}{\sum_{j=0}^M{w_j}},
\end{aligned}
\end{equation}
where $M$ is the number of all positive points in a training image.
Secondly, once a mini-batch is sampled, the training weights of each point in it (the original weight is $1$) will be re-distributed on the basis of central weights and keep the sum of weights within a mini-batch unchanged:
\begin{equation}\label{eq:re-distribution}
\begin{aligned}
q_i = \frac{w_iM'}{\sum_{j=0}^{M'}{w_j}},
\end{aligned}
\end{equation}
where $M'$ is the number of positive points within the current mini-batch and $q_i$ is the training weight of the current point.
$q_i$ is applied on all positive points in all training tasks.
Note that for negative points, $q_i=1$ in the classification task.

Consequently, the nearer a positive point is to the text center, the more frequently and significantly it would be involved in the training tasks.
Different from centerness proposed in FCOS, we re-weight different points in the training stage and no affiliated scores need to be predicted during testing.

\paragraph{\textbf{Loss function.}}
The TextRay adopts two-branch multi-task joint optimization.
For the text/non-text classification branch, the label $o^*$ for positive/negative points are set to 1/0 respectively.
For the geometric encodings regression branch, only positive points are involved.
The overall loss function of the TextRay is as follows:
\begin{equation}\label{eq:loss}
\begin{aligned}
L = \frac{1}{N_{cls}} &\sum_{i}^{N_{cls}} q_i L_{cls}(o_i, {o_i}^*) + \\
 \frac{1}{N_{reg}} &\sum_{j}^{N_{reg}} q_j L_{content}(\bm{c}_j, {\bm{c}_j}^*)
+ \frac{1}{N_{reg}} \sum_{j}^{N_{reg}} q_j L_{reg}(\bm{t}_j, {\bm{t}_j}^*).
\end{aligned}
\end{equation}
where $L_{cls}$is the classification loss, $L_{content}$ and $L_{reg}$ are losses for geometric encodings.
$\bm{t}_j=[s_j, \Delta x_j, \Delta y_j]$ represents the scale and center point offset targets of positive point $j$.
$\bm{c}_j$ is the predicted shape vector of of the $j$-th positive point.
$o_i$ is the predicted probability of the $i$-th point being text region.
${o_i}^*$, ${\bm{c}_j}^*$ and ${\bm{t}_j}^*$ are the ground-truth values.

Similar to RPN~\cite{faster-rcnn}, $L_{cls}$ is a two-class softmax loss for classification task.
$L_{content}$ is the content loss introduced in Section~\ref{sec:content}, and $L_{reg}$ is the smooth-$L_1$ loss.
These three terms are normalized by $N_{cls}$ and $N_{reg}$, where $N_{reg}$ is the number of positive points and $N_{cls}$ is the number of positive and negative points combined.

\section{Experiments}
\begin{figure}[t]
\begin{center}
\includegraphics[width=0.9\linewidth]{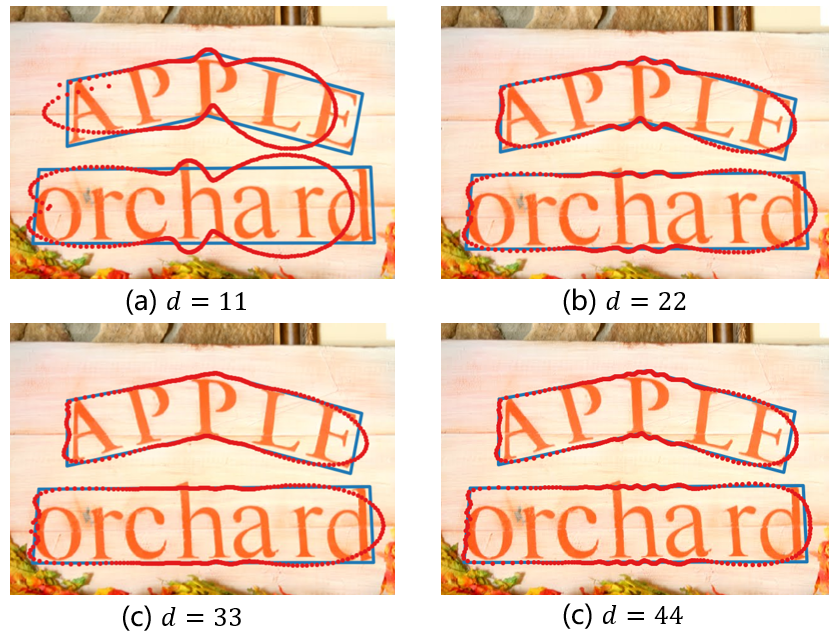}
\end{center}
\vspace{-1.0em}
   \caption{Visualization of reconstructed contours (red) from 11, 22, 33 and 44-degree Chebyshev polynomials fitting and ground-truth polygons (blue). As shown in the figures, the reconstruction error decreases with higher fitting degree.}
\label{fig:degree}
\vspace{-1.5em}
\end{figure}

\subsection{Datasets}
We evaluate the TextRay on three standard arbitrary-shaped scene text detection benchmarks: SCUT-CTW1500~\cite{ctw1500}, TotalText~\cite{totaltext} and ICDAR-ArT~\cite{ArT}.

\textit{\textbf{SCUT-CTW1500.}}
SCUT-CTW1500~\cite{ctw1500} is a challenging dataset that contains multi-oriented, curved and irregular-shaped text.
It consists of 1000 training images and 500 test images.
SCUT-CTW1500 dataset provides text line-level annotations in the form of 14 vertices.
%The text regions in SCUT-CTW1500 are given as simple polygons annotated by 14 vertices.

\textit{\textbf{TotalText.}}
TotalText~\cite{totaltext} is another curved text benchmark which consist of 1255 training images and 300 test images.
This dataset includes horizontal, multi-oriented, and curved texts.
Different from SCUT-CTW1500, the texts in TotalText are labeled at word level with adaptive number of vertices.

\textit{\textbf{ICDAR-ArT.}}
ICDAR-ArT~\cite{ArT} is a large-scale multi-lingual arbitrary-shaped text detection benchmark.
It consists of 5603 training images and 4563 test images.
Note that the whole datasets of SCUT-CTW1500 and TotalText are included in the training set of ICDAR-ArT with changes in RGB and annotation style.
The text regions are labeled with polygons annotated by adaptive number of vertices.

We adopt the same standard evaluation metric as the ICDAR challenges for all experiments.

\subsection{Implementation Details}
\label{sec:implement}
Our TextRay is implemented on PyTorch~\cite{pytorch} and MMDetection~\cite{mmdetection}.
We use ResNet50~\cite{resnet} pre-trained on ImageNet~\cite{ImageNet} as our backbone and adopt the FPN~\cite{fpn} neck for multi-level feature maps.
For the training on SCUT-CTW1500 and TotalText, we employ $\{P_3, P_4, P_5, P_6\}$ of the FPN whose resolutions scale from $8\times8$ to $64\times64$ down-sampling of the input image.
And the according relative size ranges of multi-level prediction are $[0, 0.3]$, $[0.2, 0.55]$, $[0.45, 0.8]$ and $[0.7, \infty]$.
For the training on ICDAR-ArT, we add an extra $P_7$ whose resolution is $128\times128$ down-sampling of the input image, for the scales of samples in ICDAR-ArT change more drastically.
And the multi-level ranges are set as $[0, 0.25]$, $[0.15, 0.45]$, $[0.35, 0.65]$, $[0.55, 0.85]$ and $[0.75, \infty]$.

The TextRay is optimized end-to-end by using stochastic gradient descent (SGD) with a momentum of $0.9$ and weight decay of $0.0001$.
We adopt the ``cosine'' policy to adjust the learning rate which starts at $0.08$.
An online data augmentation strategy is adopted: first we conduct random color jitter, then we randomly crop a square region on each image, and resize the cropped image to $960\times960$.
The consequent rescale factors with respect to the origin resolution are limited within ranges $[0.5, 4]$, $[0.5, 5]$ and $[0.5, 6]$ for ICDAR-ArT, SCUT-CTW1500 and TotalText, respectively.
Regions cropped out of the image boundary will be padded with zeros.
We train the models for $300$ epochs on ICDAR-ArT and $500$ epochs on SCUT-CTW1500 and TotalText with batch size $24$ on $4$ Titan X (Pascal).
During training, the negetive/positive ratio of sampling ranges from $1$ to $3$.
In the test stage, for each predicted geometric encoding we reconstruct $360$ points on the contour and uniformly sample $36$ points according to the perimeter as the final polygon detection.

\subsection{Evaluation on SCUT-CTW1500}

\begin{figure*}
\begin{center}
\includegraphics[width=1.0\linewidth]{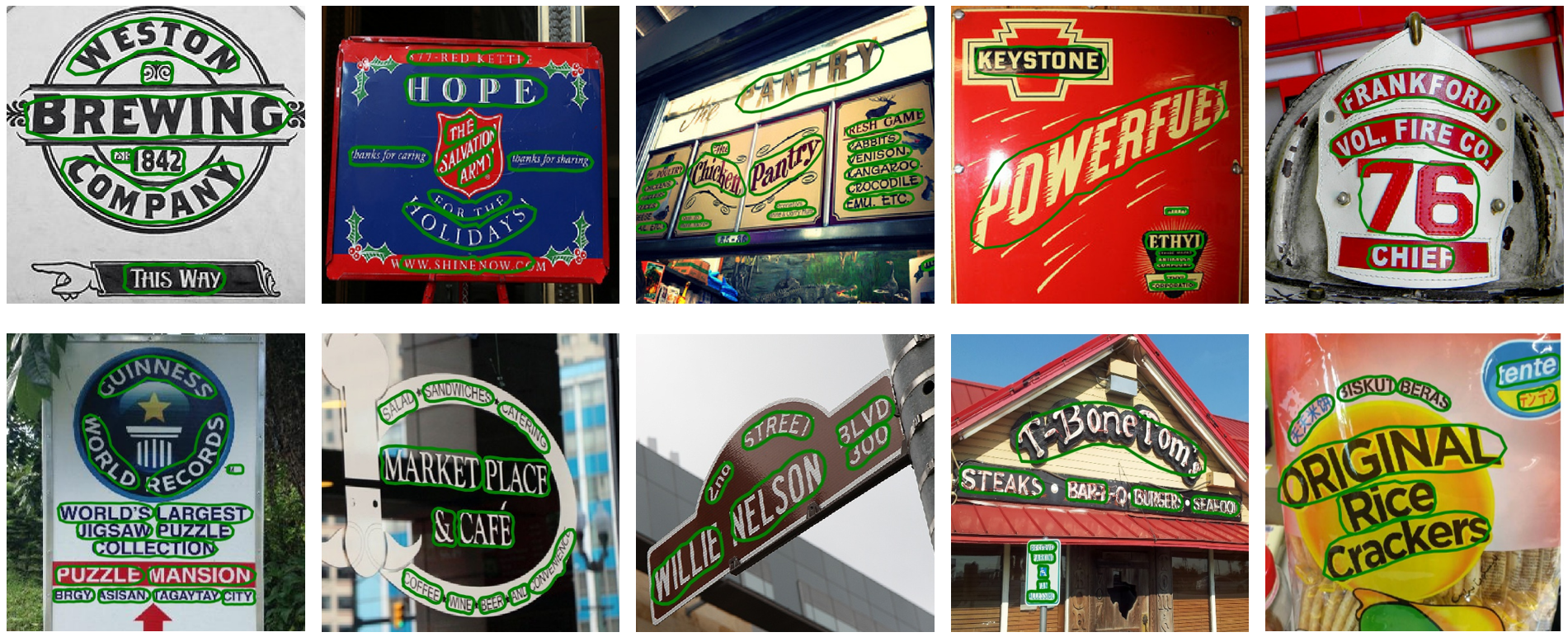}
\end{center}
\vspace{-1.0em}
   \caption{Qualitative results on SCUT-CTW1500 (first line) and TotalText (second line). The 36-point detections are visualized as green polygons.}
\label{fig:results}
\vspace{-0.5em}
\end{figure*}

We evaluate our TextRay on SCUT-CTW1500 which contains challenging rotated, curve and long line-level instances.
The fitting degree of Chebyshev polynomials is set to $K=44$ in this case.
For the lack of arbitrary-shaped text dataset, we conduct pre-training on a selected subset from ICDAR-ArT by excluding the test set of SCUT-CTW1500.
We pre-train on the selected dataset for $300$ epochs and fine-tune on SCUT-CTW1500 for $500$ epochs.
The initial learning rate of fine-tuning stage is set to $0.01$.
To deal with hard negatives, we regard points on feature maps with central weights smaller than $0.1$ as negative samples, those from $0.1$ to $0.4$ as ignored samples and those with central weights over $0.4$ as positive samples.
During testing, we resize the test images to the maximum size of $800\times640$, keeping the aspect ratios unchanged.
We take detections whose predicted scores are over $0.95$ as final results.

\begin{table}[t]% TODO: check if the comparison is extensive
	\caption{Results on SCUT-CTW1500.}
	\vspace{-1.0em}
	\centering
	\small
	\label{tb4:ctw}
	\begin{tabular}{ c||c c c }
		\hline
		\textbf{Method} & \textbf{Precision} & \textbf{Recall} & \textbf{F-measure}\tabularnewline
		\hline
		TextSnake~\cite{textsnake} &  67.90  &  \textbf{85.30} &  75.60  \tabularnewline
		
		LOMO~\cite{zhang2019look} & \textbf{89.20}  & 69.60  & 78.40 \tabularnewline
		
		CSE~\cite{towardsrobust} & 81.10  & 76.00  & 78.40 \tabularnewline
		
        PSENet-4s~\cite{wang2019shape} &  82.09  &  77.84 &  79.90  \tabularnewline
		
		Wang \textit{et al.}~\cite{wang2019arbitrary} & 80.10  & 80.20  & 80.10 \tabularnewline

		SAST~\cite{wang2019single} & 85.31  & 77.05  & 80.97 \tabularnewline

        TextField~\cite{textfield} & 83.00  & 79.80  & 81.40 \tabularnewline
		\hline
        \textbf{TextRay} & 82.80  & 80.35  & \textbf{81.56} \tabularnewline
		\hline
	\end{tabular}
    \vspace{-0.5em}
\end{table}

\begin{table}[t]% TODO: check if the comparison is extensive
	\caption{Results on TotalText.}
	\vspace{-1.0em}
	\centering
	\small
	\label{tb5:total}
	\begin{tabular}{ c||c c c }
		\hline
		\textbf{Method} & \textbf{Precision} & \textbf{Recall} & \textbf{F-measure}\tabularnewline
		\hline
		TextSnake~\cite{textsnake} &  82.70  &  74.50 &  78.40  \tabularnewline
		
		Wang \textit{et al.}~\cite{wang2019arbitrary} & 80.90  & 76.20  & 78.50 \tabularnewline
		
		PSENet-4s~\cite{wang2019shape} & 85.54  & 75.23  & 79.61 \tabularnewline
		
        %PSENet-1s~\cite{wang2019shape} &  84.02  &  77.96 &  80.87  \tabularnewline
		
		SAST~\cite{wang2019single} & 83.77  & 76.86  & 80.17 \tabularnewline
		
		CSE~\cite{towardsrobust} & 81.40  & 79.10  & 80.20 \tabularnewline

        TextDragon~\cite{feng2019textdragon} & \textbf{85.60}  & 75.70  & 80.30 \tabularnewline

        TextField~\cite{textfield} & 81.20  & \textbf{79.90}  & \textbf{80.60} \tabularnewline
		\hline
        \textbf{TextRay} & 83.49  & 77.88  & 80.59 \tabularnewline
		\hline
	\end{tabular}
    \vspace{-1.0em}
\end{table}

Quantitative results are shown in Table~\ref{tb4:ctw}.
As a light-weighted one-stage regression-based detector, the TextRay outperforms many segmentation-based and two-stage detectors according to the statistics and achieve competitive results on SCUT-CTW1500.
The qualitative results of SCUT-CTW1500 dataset is shown in the first line of Figure~\ref{fig:results}.
As we can see, the TextRay is able to accurately detect rotated, curve, irregular and long text lines.

\begin{table*}[t]% TODO: check if the comparison is extensive
	\caption{Results of different modeling methods and fitting degrees on ICDAR-ArT, SCUT-CTW1500 and TotalText.}
	\vspace{-1.0em}
	\centering
	\small
	\label{tbl:degrees}
	\begin{tabular}{|c|c|c|c|c|c|c|c|c|c|}
		\hline
         & \multicolumn{3}{c|}{\textbf{ICDAR-ArT}} & \multicolumn{3}{|c|}{\textbf{SCUT-CTW1500}} & \multicolumn{3}{c|}{\textbf{TotalText}}  \tabularnewline
        \hline \textbf{Method}&\textbf{Precision}&\textbf{Recall}&\textbf{F-measure}&\textbf{Precision}&\textbf{Recall}&\textbf{F-measure}&\textbf{Precision}&\textbf{Recall}&\textbf{F-measure}\tabularnewline
		\hline
		TextRay\_Cartesian  &  71.77       &  51.71       &  60.11       &  78.70       &  75.52       &  77.08       &  78.39       &  71.11       &  74.57       \tabularnewline
		\hline
		TextRay\_360r       &\textbf{78.24}&  57.11       &  66.03       &  79.26       &  77.09       &  78.16       &  79.58       &  75.67       &  77.57       \tabularnewline
		\hline
		TextRay\_cheby\_11d &  71.69       &  53.90       &  61.54       &  74.21       &  68.84       &  71.42       &  80.64       &  72.42       &  76.31       \tabularnewline
		\hline
        TextRay\_cheby\_22d &  75.30       &  58.22       &  65.67       &  78.95       &  76.27       &  77.59       &  79.95       &  75.26       &  77.53       \tabularnewline
		\hline
		TextRay\_cheby\_33d &  74.80       &  58.60       &  65.74       &\textbf{80.22}&  77.48       &  78.83       &\textbf{81.05}&\textbf{76.66}&\textbf{78.79}\tabularnewline
		\hline
		TextRay\_cheby\_44d &  75.97       &  58.60       &\textbf{66.17}&  79.69       &  78.16       &  78.92       &  80.96       &  75.62       &  78.20       \tabularnewline
		\hline
        TextRay\_cheby\_55d &  74.94       &\textbf{58.80}&  65.90       &  79.79       &  77.57       &  78.66       &  80.11       &  74.72       &  77.32       \tabularnewline
		\hline
		TextRay\_cheby\_66d &  75.82       &  58.60       &  66.11       &  80.19       &\textbf{78.36}&\textbf{79.26}&  80.35       &  75.30       &  77.74       \tabularnewline
		\hline
	\end{tabular}
	\vspace{-0.5em}
\end{table*}

\subsection{Evaluation on TotalText}
We also evaluate our TextRay on TotalText which contains word-level instances.
The fitting degree of Chebyshev polynomials is set to $K=33$ in this case.
Similar to SCUT-CTW1500, we exclude the test set of TotalText from ICDAR-ArT to conduct pre-training.
The training settings are the same with SCUT-CTW1500.
For the instances are at word level, to maintain a higher recall rate, we regard points with central weights equal to $0$ as negative samples, those from $0$ to $0.1$ as ignored samples and those with central weights over $0.1$ as positive samples.
In the test stage, we resize the test images to the maximum size of $960\times960$ while preserving the aspect ratios.
Detections with classification scores over $0.995$ are taken as final results.

The quantitative results are shown in Table~\ref{tb5:total}.
According to the statistics, the TextRay surpasses many state-of-the-art methods and achieves competitive results on word-level TotalText dataset.
The second line in Figure~\ref{fig:results} shows some qualitative results on TotalText.
As demonstrated, the TextRay successfully detects arbitrary-shaped text instances at word-level.

\subsection{Ablation Study}
We conduct a series of ablation studies on ICDAR-ArT, SCUT-CTW1500 and TotalText to demonstrate the effectiveness of our global geometric modeling method and the TextRay framework.
%We train the following models on these three datasets with the implementation details described in Section~\ref{sec:implement}.
All experiments in this section are conducted \textit{without pre-training}.
When testing on ICDAR-ArT, the test images are re-scaled to [1280, 960] with aspect ratio unchanged, and the negative and positive thresholds of central weights are set to $0.1$ and $0.2$.
A predicted polygon is taken as a detection if its classification score is over $0.99$.
%Details on the other two datasets are the same with previous sections.

\textit{\textbf{Influence of fitting degrees.}}
The space mapping of contours from shape space to parameter space through Chebyshev polynomials curve fitting is a process with error.
As shown in Figure~\ref{fig:degree}, the reconstruction error decreases with higher fitting degrees and thus improve the performance of our TextRay.
However, while the fitting degree increases, the error decrease slows down and the learning difficulty increases.
To study the trade-off between reconstruction error and training difficulty, we conduct a set of comparisons on fitting degree.
The third to the last lines in Table~\ref{tbl:degrees} shows that as the fitting degree going up, the performance reaches a bottleneck.
More specifically, for long line-level instances (SCUT-CTW1500), the optimal fitting degree is higher than word-level instances (TotalText).
The reason is that the larger the aspect ratio is, the larger deviation the curve would preserve, and thus more parameters are needed during curve fitting.
As a trade-off, we choose 33-degree Chebyshev polynomials for word-level TotalText and 44-degree for line-level SCUT-CTW1500 in our experiments.

\begin{figure}[t]
\begin{center}
\includegraphics[width=1.0\linewidth]{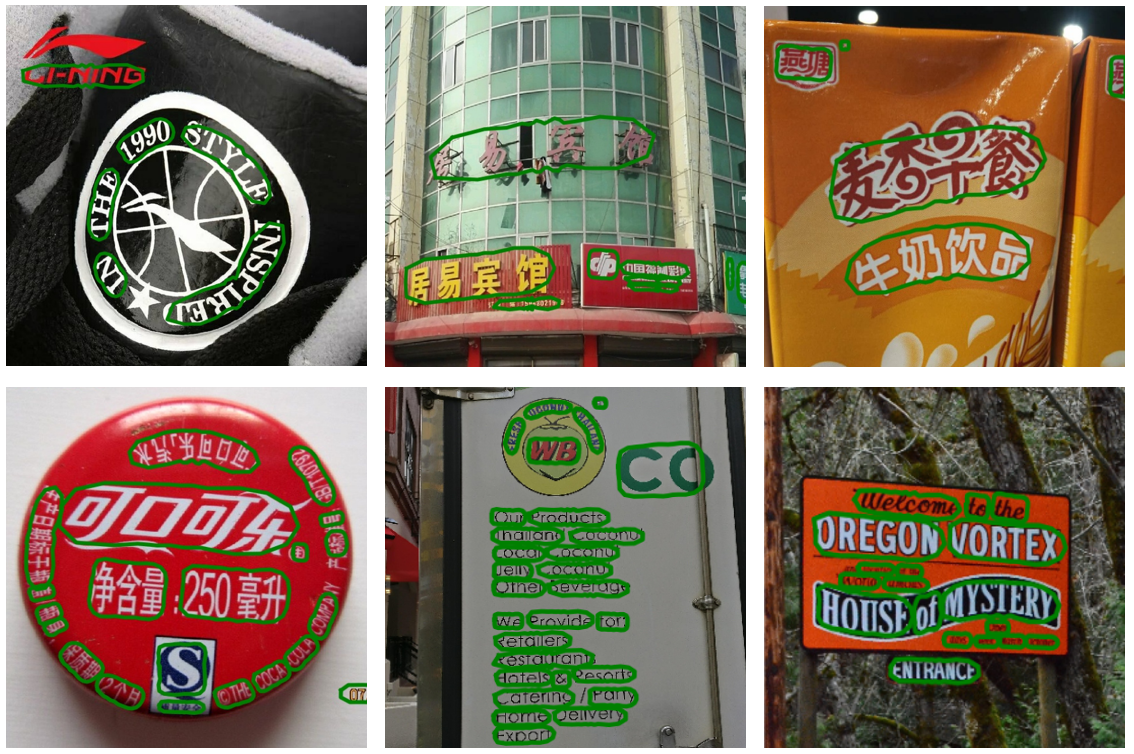}
\end{center}
\vspace{-1.0em}
   \caption{Qualitative results on ICDAR-ArT. The 36-point detections are visualized as green polygons.}
\label{fig:art}
\vspace{-1.0em}
\end{figure}

\textit{\textbf{Comparison with other global modelings.}}
To demonstrate the effectiveness of our global geometric modeling method, we conduct two comparisons.
The first one is the given global modeling for arbitrary-shaped texts in the benchmarks, that are, polygons denoted by adaptive number (ICDAR-ArT and TotalText) or fixed number (SCUT-CTW1500) of points in Cartesian coordinates.
To embed the non-unified representation into our framework, we extend all the ground-truths in ICDAR-ArT and TotalText datasets into 12 points by interpolation.
On SCUT-CTW1500 dataset we directly use its 14-point polygon annotations.
The regression targets are the offsets from the current location on the feature map to the 12 or 14 points under Cartesian system.
This experiment is denoted by \textit{TextRay\_Cartesian}.
The second comparison is to directly use the sampled 360 radiuses as the global representation of instances.
The regression targets for this experiment are 360 normalized radiuses, normalization factor $s$, and center point offset $(\Delta x, \Delta y)$.
We name this experiment \textit{TextRay\_360r}.
In these comparison experiments, independent points or radiuses are directly regressed in the TextRay framework with smooth-L1 loss under the same training settings.
Table~\ref{tbl:degrees} shows the experimental results of different modeling methods.
As we can see, the intuitive global modeling method \textit{TextRay\_Cartesian} does not perform well in arbitrary-shaped text detection.
And the \textit{TextRay\_360r} performs better against our geometric modeling methods with low fitting degrees but is surpassed by higher fitting degrees.
The comprehensive comparisons demonstrate the superiority of our modeling method.
Figure~\ref{fig:art} shows some qualitative results of $TextRay\_cheby\_44d$ on ICDAR-ArT dataset.

\begin{table}[t]% TODO: check if the comparison is extensive
	\caption{Evaluation of content loss and central-weighted training strategy on SCUT-CTW1500.}
	\vspace{-0.75em}
	\centering
	\small
	\label{tb2:techniques}
	\begin{tabular}{|c|c|c|c|c|}
		\hline
		\textbf{ContentLoss}&\textbf{CentralWeight}&\textbf{Precision}&\textbf{Recall}&\textbf{F-measure}\tabularnewline
		\hline
		  &  &  75.71  &  71.12  &  73.34  \tabularnewline
		\hline
		  $\textbf{\checkmark}$ &  &  78.97  &  76.01  &  77.46 \tabularnewline
		\hline
		  & $\textbf{\checkmark}$ &  77.45   &   75.10  &  76.25  \tabularnewline
		\hline
         $\textbf{\checkmark}$ & $\textbf{\checkmark}$ &  \textbf{79.69}  &  \textbf{78.16}  &  \textbf{78.92}  \tabularnewline
		\hline
	\end{tabular}
	\vspace{-1.25em}
\end{table}

\textit{\textbf{Effectiveness of content loss and central-weighted training.}}
To demonstrate the effectiveness of our proposed content loss and central-weighted training strategy, we carry out comparison experiments on SCUT-CTW1500 on the basis of $TextRay\_cheby\_44d$.
Except for the utilization of these two techniques, other experimental settings are all the same.
According to the quantitative results in Table~\ref{tb2:techniques}, the content loss and central-weighted training strategy independently improve the comprehensive performance (i.e., F-measure) by $4.12\%$ and $2.91\%$, respectively.
And when simultaneously employed (i.e., the proposed TextRay), the performance is improved by a large margin of $5.58\%$.

\begin{figure}[t]
\begin{center}
\includegraphics[width=1.0\linewidth]{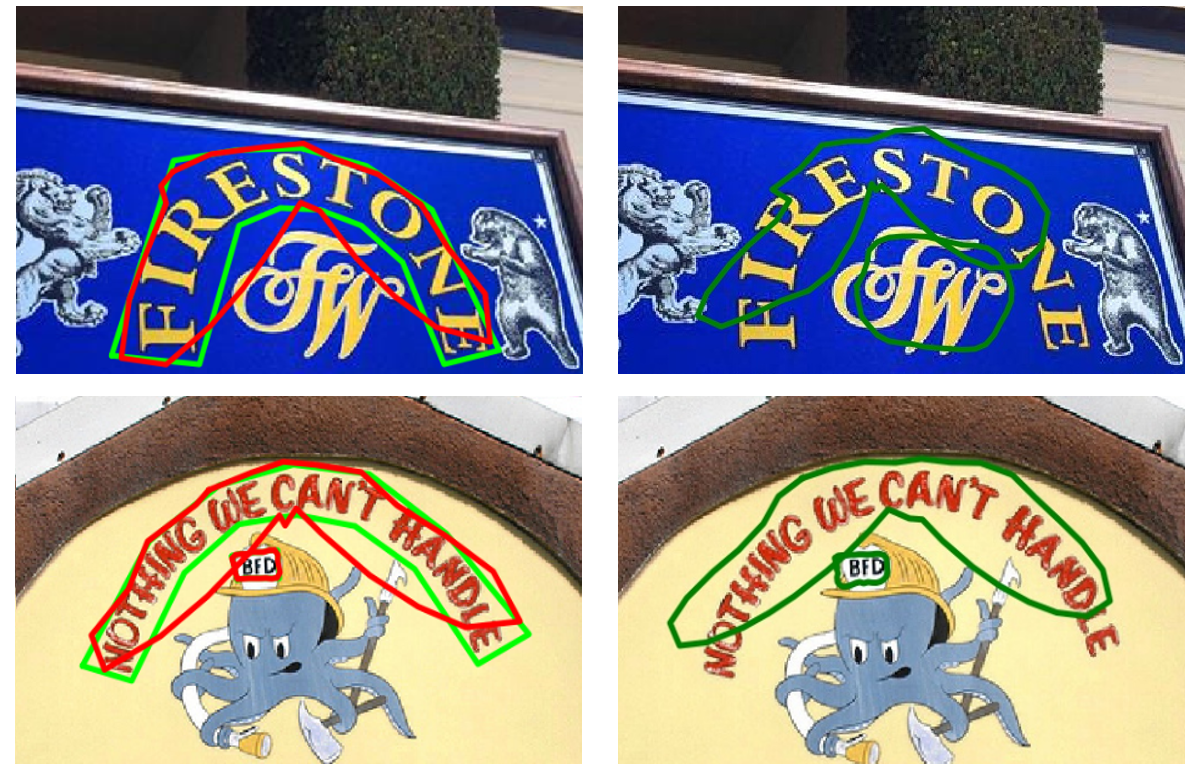}
\end{center}
\vspace{-1.0em}
   \caption{Demonstration of the limitation of our global geometric modeling method. The left column shows the ground-truth polygons (lime) and the reconstructed contours (red) from the ground-truth geometric encodings. The right column shows the detections (green). }
\label{fig:limit}
\vspace{-1.25em}
\end{figure}

\subsection{Limitation}
In this subsection we discuss the limitation of our global geometric modeling method.
Though our modeling method achieves good performances on the benchmarks, there are still some intrinsic flaws that cannot be perfectly solved.
As illustrated in the left collum of Figure~\ref{fig:limit}, for severely non-convex contours, the reconstruction error is already beyond acceptable in the geometric encoding procedure.
And thus it is impossible to successfully recover the contours under acceptable error with these geometric encodings in the decoding process.
This is because during radiuses sampling, the rays emitted from text center will frequently intersect the contour for more than once.
And each time we choose one of the intersections (i.e., the farthest one in this work), we lose a part of the contour information.
The missing of information accumulates in severely non-convex contours and makes successful reconstruction impossible.
The right column of Figure~\ref{fig:limit} shows these failed detections of our TextRay.
Though severely non-convex instances are of a very small proportion in current benchmarks, solving this limitation in our global modeling method is an important research point of our future work.

\section{Conclusion}
In this paper, we propose an arbitrary-shaped scene text detection method, namely TextRay, which conducts contour-based geometric modeling and geometric parameter learning within a single-shot anchor-free framework.
We model arbitrary shapes under polar system and represent contours with geometric encodings through bidirectional space mapping between shape space and parameter space.
The geometric encodings are effectively learned with the differentiable mapping scheme and central-weighted training strategy.
TextRay is able to accurately detect arbitrary-shaped texts and output simple polygon detections at one pass.
\footnote{
	\textbf{Acknowledgements.}
    This work is supported by key scientific technological innovation research project by Ministry of Education, Zhejiang Provincial Natural Science Foundation of China under Grant LR19F020004, Baidu AI Frontier Technology Joint Research Program, and Zhejiang University K.P.Chao's High Technology Development Foundation.
}
\bibliographystyle{ACM-Reference-Format}
\bibliography{sample-base}


\begin{thebibliography}{40}

%%% the Web address.

\ifx \showCODEN    \undefined \def \showCODEN     #1{\unskip}     \fi
\ifx \showDOI      \undefined \def \showDOI       #1{#1}\fi
\ifx \showISBNx    \undefined \def \showISBNx     #1{\unskip}     \fi
\ifx \showISBNxiii \undefined \def \showISBNxiii  #1{\unskip}     \fi
\ifx \showISSN     \undefined \def \showISSN      #1{\unskip}     \fi
\ifx \showLCCN     \undefined \def \showLCCN      #1{\unskip}     \fi
\ifx \shownote     \undefined \def \shownote      #1{#1}          \fi
\ifx \showarticletitle \undefined \def \showarticletitle #1{#1}   \fi
\ifx \showURL      \undefined \def \showURL       {\relax}        \fi
% invisible to TeX
\providecommand\bibfield[2]{#2}
\providecommand\bibinfo[2]{#2}
\providecommand\natexlab[1]{#1}
\providecommand\showeprint[2][]{arXiv:#2}

\bibitem[\protect\citeauthoryear{Baek, Lee, Han, Yun, and Lee}{Baek
  et~al\mbox{.}}{2019}]%
        {baek2019character}
\bibfield{author}{\bibinfo{person}{Youngmin Baek}, \bibinfo{person}{Bado Lee},
  \bibinfo{person}{Dongyoon Han}, \bibinfo{person}{Sangdoo Yun}, {and}
  \bibinfo{person}{Hwalsuk Lee}.} \bibinfo{year}{2019}\natexlab{}.
\newblock \showarticletitle{Character Region Awareness for Text Detection}. In
  \bibinfo{booktitle}{\emph{{IEEE} Conference on Computer Vision and Pattern
  Recognition, {CVPR}}}. \bibinfo{publisher}{{IEEE} Computer Society},
  \bibinfo{pages}{9365--9374}.
\newblock


\bibitem[\protect\citeauthoryear{Bodla, Singh, Chellappa, and Davis}{Bodla
  et~al\mbox{.}}{2017}]%
        {soft-nms}
\bibfield{author}{\bibinfo{person}{Navaneeth Bodla}, \bibinfo{person}{Bharat
  Singh}, \bibinfo{person}{Rama Chellappa}, {and} \bibinfo{person}{Larry~S.
  Davis}.} \bibinfo{year}{2017}\natexlab{}.
\newblock \showarticletitle{Soft-NMS - Improving Object Detection with One Line
  of Code}. In \bibinfo{booktitle}{\emph{{IEEE} International Conference on
  Computer Vision, {ICCV}}}. \bibinfo{publisher}{{IEEE} Computer Society},
  \bibinfo{pages}{5562--5570}.
\newblock


\bibitem[\protect\citeauthoryear{Chen, Wang, Pang, Cao, Xiong, Li, Sun, Feng,
  Liu, Xu, Zhang, Cheng, Zhu, Cheng, Zhao, Li, Lu, Zhu, Wu, Dai, Wang, Shi,
  Ouyang, Loy, and Lin}{Chen et~al\mbox{.}}{2019}]%
        {mmdetection}
\bibfield{author}{\bibinfo{person}{Kai Chen}, \bibinfo{person}{Jiaqi Wang},
  \bibinfo{person}{Jiangmiao Pang}, \bibinfo{person}{Yuhang Cao},
  \bibinfo{person}{Yu Xiong}, \bibinfo{person}{Xiaoxiao Li},
  \bibinfo{person}{Shuyang Sun}, \bibinfo{person}{Wansen Feng},
  \bibinfo{person}{Ziwei Liu}, \bibinfo{person}{Jiarui Xu},
  \bibinfo{person}{Zheng Zhang}, \bibinfo{person}{Dazhi Cheng},
  \bibinfo{person}{Chenchen Zhu}, \bibinfo{person}{Tianheng Cheng},
  \bibinfo{person}{Qijie Zhao}, \bibinfo{person}{Buyu Li}, \bibinfo{person}{Xin
  Lu}, \bibinfo{person}{Rui Zhu}, \bibinfo{person}{Yue Wu},
  \bibinfo{person}{Jifeng Dai}, \bibinfo{person}{Jingdong Wang},
  \bibinfo{person}{Jianping Shi}, \bibinfo{person}{Wanli Ouyang},
  \bibinfo{person}{Chen~Change Loy}, {and} \bibinfo{person}{Dahua Lin}.}
  \bibinfo{year}{2019}\natexlab{}.
\newblock \showarticletitle{MMDetection: Open MMLab Detection Toolbox and
  Benchmark}.
\newblock \bibinfo{journal}{\emph{CoRR}}  \bibinfo{volume}{abs/1906.07155}
  (\bibinfo{year}{2019}).
\newblock


\bibitem[\protect\citeauthoryear{Chng and Chan}{Chng and Chan}{2017}]%
        {totaltext}
\bibfield{author}{\bibinfo{person}{Chee~Kheng Chng} {and}
  \bibinfo{person}{Chee~Seng Chan}.} \bibinfo{year}{2017}\natexlab{}.
\newblock \showarticletitle{Total-Text: {A} Comprehensive Dataset for Scene
  Text Detection and Recognition}. In \bibinfo{booktitle}{\emph{14th {IAPR}
  International Conference on Document Analysis and Recognition, {ICDAR}}}.
  \bibinfo{publisher}{{IEEE} Computer Society}, \bibinfo{pages}{935--942}.
\newblock


\bibitem[\protect\citeauthoryear{Chng, Ding, Liu, Karatzas, Chan, Jin, Liu,
  Sun, Ng, Luo, Ni, Fang, Zhang, and Han}{Chng et~al\mbox{.}}{2019}]%
        {ArT}
\bibfield{author}{\bibinfo{person}{Chee~Kheng Chng}, \bibinfo{person}{Errui
  Ding}, \bibinfo{person}{Jingtuo Liu}, \bibinfo{person}{Dimosthenis Karatzas},
  \bibinfo{person}{Chee~Seng Chan}, \bibinfo{person}{Lianwen Jin},
  \bibinfo{person}{Yuliang Liu}, \bibinfo{person}{Yipeng Sun},
  \bibinfo{person}{Chun~Chet Ng}, \bibinfo{person}{Canjie Luo},
  \bibinfo{person}{Zihan Ni}, \bibinfo{person}{ChuanMing Fang},
  \bibinfo{person}{Shuaitao Zhang}, {and} \bibinfo{person}{Junyu Han}.}
  \bibinfo{year}{2019}\natexlab{}.
\newblock \showarticletitle{{ICDAR2019} Robust Reading Challenge on
  Arbitrary-Shaped Text - RRC-ArT}. In \bibinfo{booktitle}{\emph{International
  Conference on Document Analysis and Recognition, {ICDAR}}}.
  \bibinfo{publisher}{{IEEE} Computer Society}, \bibinfo{pages}{1571--1576}.
\newblock


\bibitem[\protect\citeauthoryear{Deng, Liu, Li, and Cai}{Deng
  et~al\mbox{.}}{2018}]%
        {deng2018pixellink}
\bibfield{author}{\bibinfo{person}{Dan Deng}, \bibinfo{person}{Haifeng Liu},
  \bibinfo{person}{Xuelong Li}, {and} \bibinfo{person}{Deng Cai}.}
  \bibinfo{year}{2018}\natexlab{}.
\newblock \showarticletitle{PixelLink: Detecting Scene Text via Instance
  Segmentation}. In \bibinfo{booktitle}{\emph{Proceedings of the Thirty-Second
  {AAAI} Conference on Artificial Intelligence}},
  \bibfield{editor}{\bibinfo{person}{Sheila~A. McIlraith} {and}
  \bibinfo{person}{Kilian~Q. Weinberger}} (Eds.). \bibinfo{publisher}{{AAAI}
  Press}, \bibinfo{pages}{6773--6780}.
\newblock


\bibitem[\protect\citeauthoryear{Duan, Xu, Kuang, Yue, Sun, Guan, and
  Zhang}{Duan et~al\mbox{.}}{2019}]%
        {geometry_normalize}
\bibfield{author}{\bibinfo{person}{Jiaqi Duan}, \bibinfo{person}{Youjiang Xu},
  \bibinfo{person}{Zhanghui Kuang}, \bibinfo{person}{Xiaoyu Yue},
  \bibinfo{person}{Hongbin Sun}, \bibinfo{person}{Yue Guan}, {and}
  \bibinfo{person}{Wayne Zhang}.} \bibinfo{year}{2019}\natexlab{}.
\newblock \showarticletitle{Geometry Normalization Networks for Accurate Scene
  Text Detection}. In \bibinfo{booktitle}{\emph{{IEEE} International Conference
  on Computer Vision, {ICCV}}}. \bibinfo{publisher}{{IEEE} Computer Society},
  \bibinfo{pages}{9136--9145}.
\newblock


\bibitem[\protect\citeauthoryear{Feng, He, Yin, Zhang, and Liu}{Feng
  et~al\mbox{.}}{2019}]%
        {feng2019textdragon}
\bibfield{author}{\bibinfo{person}{Wei Feng}, \bibinfo{person}{Wenhao He},
  \bibinfo{person}{Fei Yin}, \bibinfo{person}{Xu{-}Yao Zhang}, {and}
  \bibinfo{person}{Cheng{-}Lin Liu}.} \bibinfo{year}{2019}\natexlab{}.
\newblock \showarticletitle{TextDragon: An End-to-End Framework for Arbitrary
  Shaped Text Spotting}. In \bibinfo{booktitle}{\emph{{IEEE} International
  Conference on Computer Vision, {ICCV}}}. \bibinfo{publisher}{{IEEE} Computer
  Society}, \bibinfo{pages}{9075--9084}.
\newblock


\bibitem[\protect\citeauthoryear{Girshick}{Girshick}{2015}]%
        {fast-rcnn}
\bibfield{author}{\bibinfo{person}{Ross~B. Girshick}.}
  \bibinfo{year}{2015}\natexlab{}.
\newblock \showarticletitle{Fast {R-CNN}}. In \bibinfo{booktitle}{\emph{{IEEE}
  International Conference on Computer Vision, {ICCV}}}.
  \bibinfo{publisher}{{IEEE} Computer Society}, \bibinfo{pages}{1440--1448}.
\newblock


\bibitem[\protect\citeauthoryear{He, Gkioxari, Doll{\'{a}}r, and Girshick}{He
  et~al\mbox{.}}{2017}]%
        {maskrcnn}
\bibfield{author}{\bibinfo{person}{Kaiming He}, \bibinfo{person}{Georgia
  Gkioxari}, \bibinfo{person}{Piotr Doll{\'{a}}r}, {and}
  \bibinfo{person}{Ross~B. Girshick}.} \bibinfo{year}{2017}\natexlab{}.
\newblock \showarticletitle{Mask {R-CNN}}. In \bibinfo{booktitle}{\emph{{IEEE}
  International Conference on Computer Vision, {ICCV}}}.
  \bibinfo{publisher}{{IEEE} Computer Society}, \bibinfo{pages}{2980--2988}.
\newblock


\bibitem[\protect\citeauthoryear{He, Zhang, Ren, and Sun}{He
  et~al\mbox{.}}{2016}]%
        {resnet}
\bibfield{author}{\bibinfo{person}{Kaiming He}, \bibinfo{person}{Xiangyu
  Zhang}, \bibinfo{person}{Shaoqing Ren}, {and} \bibinfo{person}{Jian Sun}.}
  \bibinfo{year}{2016}\natexlab{}.
\newblock \showarticletitle{Deep Residual Learning for Image Recognition}. In
  \bibinfo{booktitle}{\emph{{IEEE} Conference on Computer Vision and Pattern
  Recognition, {CVPR}}}. \bibinfo{publisher}{{IEEE} Computer Society},
  \bibinfo{pages}{770--778}.
\newblock


\bibitem[\protect\citeauthoryear{Liao, Zhu, Shi, Xia, and Bai}{Liao
  et~al\mbox{.}}{2018}]%
        {rotationsensitive}
\bibfield{author}{\bibinfo{person}{Minghui Liao}, \bibinfo{person}{Zhen Zhu},
  \bibinfo{person}{Baoguang Shi}, \bibinfo{person}{Gui{-}Song Xia}, {and}
  \bibinfo{person}{Xiang Bai}.} \bibinfo{year}{2018}\natexlab{}.
\newblock \showarticletitle{Rotation-Sensitive Regression for Oriented Scene
  Text Detection}. In \bibinfo{booktitle}{\emph{{IEEE} Conference on Computer
  Vision and Pattern Recognition, {CVPR}}}. \bibinfo{publisher}{{IEEE} Computer
  Society}, \bibinfo{pages}{5909--5918}.
\newblock


\bibitem[\protect\citeauthoryear{Lin, Doll{\'{a}}r, Girshick, He, Hariharan,
  and Belongie}{Lin et~al\mbox{.}}{2017}]%
        {fpn}
\bibfield{author}{\bibinfo{person}{Tsung{-}Yi Lin}, \bibinfo{person}{Piotr
  Doll{\'{a}}r}, \bibinfo{person}{Ross~B. Girshick}, \bibinfo{person}{Kaiming
  He}, \bibinfo{person}{Bharath Hariharan}, {and} \bibinfo{person}{Serge~J.
  Belongie}.} \bibinfo{year}{2017}\natexlab{}.
\newblock \showarticletitle{Feature Pyramid Networks for Object Detection}. In
  \bibinfo{booktitle}{\emph{{IEEE} Conference on Computer Vision and Pattern
  Recognition, {CVPR}}}. \bibinfo{publisher}{{IEEE} Computer Society},
  \bibinfo{pages}{936--944}.
\newblock


\bibitem[\protect\citeauthoryear{Liu, Anguelov, Erhan, Szegedy, Reed, Fu, and
  Berg}{Liu et~al\mbox{.}}{2016}]%
        {SSD}
\bibfield{author}{\bibinfo{person}{Wei Liu}, \bibinfo{person}{Dragomir
  Anguelov}, \bibinfo{person}{Dumitru Erhan}, \bibinfo{person}{Christian
  Szegedy}, \bibinfo{person}{Scott~E. Reed}, \bibinfo{person}{Cheng{-}Yang Fu},
  {and} \bibinfo{person}{Alexander~C. Berg}.} \bibinfo{year}{2016}\natexlab{}.
\newblock \showarticletitle{{SSD:} Single Shot MultiBox Detector}. In
  \bibinfo{booktitle}{\emph{Computer Vision - {ECCV} 2016 - 14th European
  Conference, Proceedings, Part {I}}} \emph{(\bibinfo{series}{Lecture Notes in
  Computer Science})}, Vol.~\bibinfo{volume}{9905}.
  \bibinfo{publisher}{Springer}, \bibinfo{pages}{21--37}.
\newblock


\bibitem[\protect\citeauthoryear{Liu, Chen, Shen, He, Jin, and Wang}{Liu
  et~al\mbox{.}}{2020}]%
        {abcnet}
\bibfield{author}{\bibinfo{person}{Yuliang Liu}, \bibinfo{person}{Hao Chen},
  \bibinfo{person}{Chunhua Shen}, \bibinfo{person}{Tong He},
  \bibinfo{person}{Lianwen Jin}, {and} \bibinfo{person}{Liangwei Wang}.}
  \bibinfo{year}{2020}\natexlab{}.
\newblock \showarticletitle{ABCNet: Real-time Scene Text Spotting with Adaptive
  Bezier-Curve Network}.
\newblock \bibinfo{journal}{\emph{CoRR}}  \bibinfo{volume}{abs/2002.10200}
  (\bibinfo{year}{2020}).
\newblock


\bibitem[\protect\citeauthoryear{Liu, Jin, Zhang, Luo, and Zhang}{Liu
  et~al\mbox{.}}{2019a}]%
        {ctw1500}
\bibfield{author}{\bibinfo{person}{Yuliang Liu}, \bibinfo{person}{Lianwen Jin},
  \bibinfo{person}{Shuaitao Zhang}, \bibinfo{person}{Canjie Luo}, {and}
  \bibinfo{person}{Sheng Zhang}.} \bibinfo{year}{2019}\natexlab{a}.
\newblock \showarticletitle{Curved scene text detection via transverse and
  longitudinal sequence connection}.
\newblock \bibinfo{journal}{\emph{Pattern Recognit.}}  \bibinfo{volume}{90}
  (\bibinfo{year}{2019}), \bibinfo{pages}{337--345}.
\newblock


\bibitem[\protect\citeauthoryear{Liu, Lin, Yang, Feng, Lin, and Goh}{Liu
  et~al\mbox{.}}{2018}]%
        {liu2018learning}
\bibfield{author}{\bibinfo{person}{Zichuan Liu}, \bibinfo{person}{Guosheng
  Lin}, \bibinfo{person}{Sheng Yang}, \bibinfo{person}{Jiashi Feng},
  \bibinfo{person}{Weisi Lin}, {and} \bibinfo{person}{Wang~Ling Goh}.}
  \bibinfo{year}{2018}\natexlab{}.
\newblock \showarticletitle{Learning Markov Clustering Networks for Scene Text
  Detection}. In \bibinfo{booktitle}{\emph{{IEEE} Conference on Computer Vision
  and Pattern Recognition, {CVPR}}}. \bibinfo{publisher}{{IEEE} Computer
  Society}, \bibinfo{pages}{6936--6944}.
\newblock


\bibitem[\protect\citeauthoryear{Liu, Lin, Yang, Liu, Lin, and Goh}{Liu
  et~al\mbox{.}}{2019b}]%
        {towardsrobust}
\bibfield{author}{\bibinfo{person}{Zichuan Liu}, \bibinfo{person}{Guosheng
  Lin}, \bibinfo{person}{Sheng Yang}, \bibinfo{person}{Fayao Liu},
  \bibinfo{person}{Weisi Lin}, {and} \bibinfo{person}{Wang~Ling Goh}.}
  \bibinfo{year}{2019}\natexlab{b}.
\newblock \showarticletitle{Towards Robust Curve Text Detection With
  Conditional Spatial Expansion}. In \bibinfo{booktitle}{\emph{{IEEE}
  Conference on Computer Vision and Pattern Recognition, {CVPR}}}.
  \bibinfo{publisher}{{IEEE} Computer Society}, \bibinfo{pages}{7269--7278}.
\newblock


\bibitem[\protect\citeauthoryear{Long, Shelhamer, and Darrell}{Long
  et~al\mbox{.}}{2015}]%
        {fcn}
\bibfield{author}{\bibinfo{person}{Jonathan Long}, \bibinfo{person}{Evan
  Shelhamer}, {and} \bibinfo{person}{Trevor Darrell}.}
  \bibinfo{year}{2015}\natexlab{}.
\newblock \showarticletitle{Fully convolutional networks for semantic
  segmentation}. In \bibinfo{booktitle}{\emph{{IEEE} Conference on Computer
  Vision and Pattern Recognition, {CVPR}}}. \bibinfo{publisher}{{IEEE} Computer
  Society}, \bibinfo{pages}{3431--3440}.
\newblock


\bibitem[\protect\citeauthoryear{Long, Ruan, Zhang, He, Wu, and Yao}{Long
  et~al\mbox{.}}{2018}]%
        {textsnake}
\bibfield{author}{\bibinfo{person}{Shangbang Long}, \bibinfo{person}{Jiaqiang
  Ruan}, \bibinfo{person}{Wenjie Zhang}, \bibinfo{person}{Xin He},
  \bibinfo{person}{Wenhao Wu}, {and} \bibinfo{person}{Cong Yao}.}
  \bibinfo{year}{2018}\natexlab{}.
\newblock \showarticletitle{TextSnake: {A} Flexible Representation for
  Detecting Text of Arbitrary Shapes}. In \bibinfo{booktitle}{\emph{Computer
  Vision - {ECCV} 2018 - 15th European Conference, Proceedings, Part {II}}},
  Vol.~\bibinfo{volume}{11206}. \bibinfo{publisher}{Springer},
  \bibinfo{pages}{19--35}.
\newblock


\bibitem[\protect\citeauthoryear{Lyu, Liao, Yao, Wu, and Bai}{Lyu
  et~al\mbox{.}}{2018a}]%
        {masktextspotter}
\bibfield{author}{\bibinfo{person}{Pengyuan Lyu}, \bibinfo{person}{Minghui
  Liao}, \bibinfo{person}{Cong Yao}, \bibinfo{person}{Wenhao Wu}, {and}
  \bibinfo{person}{Xiang Bai}.} \bibinfo{year}{2018}\natexlab{a}.
\newblock \showarticletitle{Mask TextSpotter: An End-to-End Trainable Neural
  Network for Spotting Text with Arbitrary Shapes}. In
  \bibinfo{booktitle}{\emph{Computer Vision - {ECCV} 2018 - 15th European
  Conference, Proceedings, Part {XIV}}},
  \bibfield{editor}{\bibinfo{person}{Vittorio Ferrari},
  \bibinfo{person}{Martial Hebert}, \bibinfo{person}{Cristian Sminchisescu},
  {and} \bibinfo{person}{Yair Weiss}} (Eds.), Vol.~\bibinfo{volume}{11218}.
  \bibinfo{publisher}{Springer}, \bibinfo{pages}{71--88}.
\newblock


\bibitem[\protect\citeauthoryear{Lyu, Yao, Wu, Yan, and Bai}{Lyu
  et~al\mbox{.}}{2018b}]%
        {cornerregion}
\bibfield{author}{\bibinfo{person}{Pengyuan Lyu}, \bibinfo{person}{Cong Yao},
  \bibinfo{person}{Wenhao Wu}, \bibinfo{person}{Shuicheng Yan}, {and}
  \bibinfo{person}{Xiang Bai}.} \bibinfo{year}{2018}\natexlab{b}.
\newblock \showarticletitle{Multi-Oriented Scene Text Detection via Corner
  Localization and Region Segmentation}. In \bibinfo{booktitle}{\emph{{IEEE}
  Conference on Computer Vision and Pattern Recognition, {CVPR}}}.
  \bibinfo{publisher}{{IEEE} Computer Society}, \bibinfo{pages}{7553--7563}.
\newblock


\bibitem[\protect\citeauthoryear{Pan, Shi, Luo, Wang, and Tang}{Pan
  et~al\mbox{.}}{2018}]%
        {dulr}
\bibfield{author}{\bibinfo{person}{Xingang Pan}, \bibinfo{person}{Jianping
  Shi}, \bibinfo{person}{Ping Luo}, \bibinfo{person}{Xiaogang Wang}, {and}
  \bibinfo{person}{Xiaoou Tang}.} \bibinfo{year}{2018}\natexlab{}.
\newblock \showarticletitle{Spatial as Deep: Spatial {CNN} for Traffic Scene
  Understanding}. In \bibinfo{booktitle}{\emph{Proceedings of the Thirty-Second
  {AAAI} Conference on Artificial Intelligence}},
  \bibfield{editor}{\bibinfo{person}{Sheila~A. McIlraith} {and}
  \bibinfo{person}{Kilian~Q. Weinberger}} (Eds.). \bibinfo{publisher}{{AAAI}
  Press}, \bibinfo{pages}{7276--7283}.
\newblock


\bibitem[\protect\citeauthoryear{Paszke, Gross, Massa, Lerer, Bradbury, Chanan,
  Killeen, Lin, Gimelshein, Antiga, Desmaison, K{\"{o}}pf, Yang, DeVito,
  Raison, Tejani, Chilamkurthy, Steiner, Fang, Bai, and Chintala}{Paszke
  et~al\mbox{.}}{2019}]%
        {pytorch}
\bibfield{author}{\bibinfo{person}{Adam Paszke}, \bibinfo{person}{Sam Gross},
  \bibinfo{person}{Francisco Massa}, \bibinfo{person}{Adam Lerer},
  \bibinfo{person}{James Bradbury}, \bibinfo{person}{Gregory Chanan},
  \bibinfo{person}{Trevor Killeen}, \bibinfo{person}{Zeming Lin},
  \bibinfo{person}{Natalia Gimelshein}, \bibinfo{person}{Luca Antiga},
  \bibinfo{person}{Alban Desmaison}, \bibinfo{person}{Andreas K{\"{o}}pf},
  \bibinfo{person}{Edward Yang}, \bibinfo{person}{Zachary DeVito},
  \bibinfo{person}{Martin Raison}, \bibinfo{person}{Alykhan Tejani},
  \bibinfo{person}{Sasank Chilamkurthy}, \bibinfo{person}{Benoit Steiner},
  \bibinfo{person}{Lu Fang}, \bibinfo{person}{Junjie Bai}, {and}
  \bibinfo{person}{Soumith Chintala}.} \bibinfo{year}{2019}\natexlab{}.
\newblock \showarticletitle{PyTorch: An Imperative Style, High-Performance Deep
  Learning Library}. In \bibinfo{booktitle}{\emph{Advances in Neural
  Information Processing Systems}}. \bibinfo{pages}{8024--8035}.
\newblock


\bibitem[\protect\citeauthoryear{Ren, He, Girshick, and Sun}{Ren
  et~al\mbox{.}}{2015}]%
        {faster-rcnn}
\bibfield{author}{\bibinfo{person}{Shaoqing Ren}, \bibinfo{person}{Kaiming He},
  \bibinfo{person}{Ross~B. Girshick}, {and} \bibinfo{person}{Jian Sun}.}
  \bibinfo{year}{2015}\natexlab{}.
\newblock \showarticletitle{Faster {R-CNN:} Towards Real-Time Object Detection
  with Region Proposal Networks}. In \bibinfo{booktitle}{\emph{Advances in
  Neural Information Processing Systems 28: Annual Conference on Neural
  Information Processing Systems, {NIPS}}}. \bibinfo{pages}{91--99}.
\newblock


\bibitem[\protect\citeauthoryear{Russakovsky, Deng, Su, Krause, Satheesh, Ma,
  Huang, Karpathy, Khosla, Bernstein, Berg, and Li}{Russakovsky
  et~al\mbox{.}}{2015}]%
        {ImageNet}
\bibfield{author}{\bibinfo{person}{Olga Russakovsky}, \bibinfo{person}{Jia
  Deng}, \bibinfo{person}{Hao Su}, \bibinfo{person}{Jonathan Krause},
  \bibinfo{person}{Sanjeev Satheesh}, \bibinfo{person}{Sean Ma},
  \bibinfo{person}{Zhiheng Huang}, \bibinfo{person}{Andrej Karpathy},
  \bibinfo{person}{Aditya Khosla}, \bibinfo{person}{Michael~S. Bernstein},
  \bibinfo{person}{Alexander~C. Berg}, {and} \bibinfo{person}{Fei{-}Fei Li}.}
  \bibinfo{year}{2015}\natexlab{}.
\newblock \showarticletitle{ImageNet Large Scale Visual Recognition Challenge}.
\newblock \bibinfo{journal}{\emph{Int. J. Comput. Vis.}} \bibinfo{volume}{115},
  \bibinfo{number}{3} (\bibinfo{year}{2015}), \bibinfo{pages}{211--252}.
\newblock


\bibitem[\protect\citeauthoryear{Tian, Shen, Chen, and He}{Tian
  et~al\mbox{.}}{2019}]%
        {fcos}
\bibfield{author}{\bibinfo{person}{Zhi Tian}, \bibinfo{person}{Chunhua Shen},
  \bibinfo{person}{Hao Chen}, {and} \bibinfo{person}{Tong He}.}
  \bibinfo{year}{2019}\natexlab{}.
\newblock \showarticletitle{{FCOS:} Fully Convolutional One-Stage Object
  Detection}. In \bibinfo{booktitle}{\emph{{IEEE} International Conference on
  Computer Vision, {ICCV}}}. \bibinfo{publisher}{{IEEE} Computer Society},
  \bibinfo{pages}{9626--9635}.
\newblock


\bibitem[\protect\citeauthoryear{Wang, Zhao, Li, Wang, and Tao}{Wang
  et~al\mbox{.}}{2018}]%
        {itn}
\bibfield{author}{\bibinfo{person}{Fangfang Wang}, \bibinfo{person}{Liming
  Zhao}, \bibinfo{person}{Xi Li}, \bibinfo{person}{Xinchao Wang}, {and}
  \bibinfo{person}{Dacheng Tao}.} \bibinfo{year}{2018}\natexlab{}.
\newblock \showarticletitle{Geometry-Aware Scene Text Detection With Instance
  Transformation Network}. In \bibinfo{booktitle}{\emph{{IEEE} Conference on
  Computer Vision and Pattern Recognition, {CVPR}}}. \bibinfo{publisher}{{IEEE}
  Computer Society}, \bibinfo{pages}{1381--1389}.
\newblock


\bibitem[\protect\citeauthoryear{Wang, Zhang, Qi, Huang, En, Han, Liu, Ding,
  and Shi}{Wang et~al\mbox{.}}{2019d}]%
        {wang2019single}
\bibfield{author}{\bibinfo{person}{Pengfei Wang}, \bibinfo{person}{Chengquan
  Zhang}, \bibinfo{person}{Fei Qi}, \bibinfo{person}{Zuming Huang},
  \bibinfo{person}{Mengyi En}, \bibinfo{person}{Junyu Han},
  \bibinfo{person}{Jingtuo Liu}, \bibinfo{person}{Errui Ding}, {and}
  \bibinfo{person}{Guangming Shi}.} \bibinfo{year}{2019}\natexlab{d}.
\newblock \showarticletitle{A Single-Shot Arbitrarily-Shaped Text Detector
  based on Context Attended Multi-Task Learning}. In
  \bibinfo{booktitle}{\emph{Proceedings of the 27th {ACM} International
  Conference on Multimedia, {MM}}}. \bibinfo{publisher}{{ACM}},
  \bibinfo{pages}{1277--1285}.
\newblock


\bibitem[\protect\citeauthoryear{Wang, Xie, Li, Hou, Lu, Yu, and Shao}{Wang
  et~al\mbox{.}}{2019b}]%
        {wang2019shape}
\bibfield{author}{\bibinfo{person}{Wenhai Wang}, \bibinfo{person}{Enze Xie},
  \bibinfo{person}{Xiang Li}, \bibinfo{person}{Wenbo Hou},
  \bibinfo{person}{Tong Lu}, \bibinfo{person}{Gang Yu}, {and}
  \bibinfo{person}{Shuai Shao}.} \bibinfo{year}{2019}\natexlab{b}.
\newblock \showarticletitle{Shape Robust Text Detection With Progressive Scale
  Expansion Network}. In \bibinfo{booktitle}{\emph{{IEEE} Conference on
  Computer Vision and Pattern Recognition, {CVPR}}}. \bibinfo{publisher}{{IEEE}
  Computer Society}, \bibinfo{pages}{9336--9345}.
\newblock


\bibitem[\protect\citeauthoryear{Wang, Xie, Song, Zang, Wang, Lu, Yu, and
  Shen}{Wang et~al\mbox{.}}{2019c}]%
        {wang2019efficient}
\bibfield{author}{\bibinfo{person}{Wenhai Wang}, \bibinfo{person}{Enze Xie},
  \bibinfo{person}{Xiaoge Song}, \bibinfo{person}{Yuhang Zang},
  \bibinfo{person}{Wenjia Wang}, \bibinfo{person}{Tong Lu},
  \bibinfo{person}{Gang Yu}, {and} \bibinfo{person}{Chunhua Shen}.}
  \bibinfo{year}{2019}\natexlab{c}.
\newblock \showarticletitle{Efficient and Accurate Arbitrary-Shaped Text
  Detection With Pixel Aggregation Network}. In
  \bibinfo{booktitle}{\emph{{IEEE} International Conference on Computer Vision,
  {ICCV}}}. \bibinfo{publisher}{{IEEE} Computer Society},
  \bibinfo{pages}{8439--8448}.
\newblock


\bibitem[\protect\citeauthoryear{Wang, Jiang, Luo, Liu, Choi, and Kim}{Wang
  et~al\mbox{.}}{2019a}]%
        {wang2019arbitrary}
\bibfield{author}{\bibinfo{person}{Xiaobing Wang}, \bibinfo{person}{Yingying
  Jiang}, \bibinfo{person}{Zhenbo Luo}, \bibinfo{person}{Cheng{-}Lin Liu},
  \bibinfo{person}{Hyunsoo Choi}, {and} \bibinfo{person}{Sungjin Kim}.}
  \bibinfo{year}{2019}\natexlab{a}.
\newblock \showarticletitle{Arbitrary Shape Scene Text Detection With Adaptive
  Text Region Representation}. In \bibinfo{booktitle}{\emph{{IEEE} Conference
  on Computer Vision and Pattern Recognition, {CVPR}}}.
  \bibinfo{publisher}{{IEEE} Computer Society}, \bibinfo{pages}{6449--6458}.
\newblock


\bibitem[\protect\citeauthoryear{Wang, Xie, Zha, Xing, Fu, and Zhang}{Wang
  et~al\mbox{.}}{2020}]%
        {contournet}
\bibfield{author}{\bibinfo{person}{Yuxin Wang}, \bibinfo{person}{Hongtao Xie},
  \bibinfo{person}{Zhengjun Zha}, \bibinfo{person}{Mengting Xing},
  \bibinfo{person}{Zilong Fu}, {and} \bibinfo{person}{Yongdong Zhang}.}
  \bibinfo{year}{2020}\natexlab{}.
\newblock \showarticletitle{ContourNet: Taking a Further Step toward Accurate
  Arbitrary-shaped Scene Text Detection}.
\newblock \bibinfo{journal}{\emph{CoRR}}  \bibinfo{volume}{abs/2004.04940}
  (\bibinfo{year}{2020}).
\newblock


\bibitem[\protect\citeauthoryear{Xie, Sun, Song, Wang, Liu, Liang, Shen, and
  Luo}{Xie et~al\mbox{.}}{2019}]%
        {xie2019polarmask}
\bibfield{author}{\bibinfo{person}{Enze Xie}, \bibinfo{person}{Peize Sun},
  \bibinfo{person}{Xiaoge Song}, \bibinfo{person}{Wenhai Wang},
  \bibinfo{person}{Xuebo Liu}, \bibinfo{person}{Ding Liang},
  \bibinfo{person}{Chunhua Shen}, {and} \bibinfo{person}{Ping Luo}.}
  \bibinfo{year}{2019}\natexlab{}.
\newblock \showarticletitle{PolarMask: Single Shot Instance Segmentation with
  Polar Representation}.
\newblock \bibinfo{journal}{\emph{CoRR}}  \bibinfo{volume}{abs/1909.13226}
  (\bibinfo{year}{2019}).
\newblock


\bibitem[\protect\citeauthoryear{Xu, Wang, Qi, and Lu}{Xu
  et~al\mbox{.}}{2019a}]%
        {xu2019explicit}
\bibfield{author}{\bibinfo{person}{Wenqiang Xu}, \bibinfo{person}{Haiyang
  Wang}, \bibinfo{person}{Fubo Qi}, {and} \bibinfo{person}{Cewu Lu}.}
  \bibinfo{year}{2019}\natexlab{a}.
\newblock \showarticletitle{Explicit Shape Encoding for Real-Time Instance
  Segmentation}. In \bibinfo{booktitle}{\emph{{IEEE} International Conference
  on Computer Vision, {ICCV}}}. \bibinfo{publisher}{{IEEE} Computer Society},
  \bibinfo{pages}{5167--5176}.
\newblock


\bibitem[\protect\citeauthoryear{Xu, Wang, Zhou, Wang, Yang, and Bai}{Xu
  et~al\mbox{.}}{2019b}]%
        {textfield}
\bibfield{author}{\bibinfo{person}{Yongchao Xu}, \bibinfo{person}{Yukang Wang},
  \bibinfo{person}{Wei Zhou}, \bibinfo{person}{Yongpan Wang},
  \bibinfo{person}{Zhibo Yang}, {and} \bibinfo{person}{Xiang Bai}.}
  \bibinfo{year}{2019}\natexlab{b}.
\newblock \showarticletitle{TextField: Learning a Deep Direction Field for
  Irregular Scene Text Detection}.
\newblock \bibinfo{journal}{\emph{{IEEE} Trans. Image Processing}}
  \bibinfo{volume}{28}, \bibinfo{number}{11} (\bibinfo{year}{2019}),
  \bibinfo{pages}{5566--5579}.
\newblock


\bibitem[\protect\citeauthoryear{Yang, Xu, Xue, Zhang, Urtasun, Wang, Lin, and
  Hu}{Yang et~al\mbox{.}}{2019}]%
        {desrep}
\bibfield{author}{\bibinfo{person}{Ze Yang}, \bibinfo{person}{Yinghao Xu},
  \bibinfo{person}{Han Xue}, \bibinfo{person}{Zheng Zhang},
  \bibinfo{person}{Raquel Urtasun}, \bibinfo{person}{Liwei Wang},
  \bibinfo{person}{Stephen Lin}, {and} \bibinfo{person}{Han Hu}.}
  \bibinfo{year}{2019}\natexlab{}.
\newblock \showarticletitle{Dense RepPoints: Representing Visual Objects with
  Dense Point Sets}.
\newblock \bibinfo{journal}{\emph{CoRR}}  \bibinfo{volume}{abs/1912.11473}
  (\bibinfo{year}{2019}).
\newblock


\bibitem[\protect\citeauthoryear{Yao, Bai, Sang, Zhou, Zhou, and Cao}{Yao
  et~al\mbox{.}}{2016}]%
        {multi-channel}
\bibfield{author}{\bibinfo{person}{Cong Yao}, \bibinfo{person}{Xiang Bai},
  \bibinfo{person}{Nong Sang}, \bibinfo{person}{Xinyu Zhou},
  \bibinfo{person}{Shuchang Zhou}, {and} \bibinfo{person}{Zhimin Cao}.}
  \bibinfo{year}{2016}\natexlab{}.
\newblock \showarticletitle{Scene Text Detection via Holistic, Multi-Channel
  Prediction}.
\newblock \bibinfo{journal}{\emph{CoRR}}  \bibinfo{volume}{abs/1606.09002}
  (\bibinfo{year}{2016}).
\newblock


\bibitem[\protect\citeauthoryear{Zhang, Liang, Huang, En, Han, Ding, and
  Ding}{Zhang et~al\mbox{.}}{2019}]%
        {zhang2019look}
\bibfield{author}{\bibinfo{person}{Chengquan Zhang}, \bibinfo{person}{Borong
  Liang}, \bibinfo{person}{Zuming Huang}, \bibinfo{person}{Mengyi En},
  \bibinfo{person}{Junyu Han}, \bibinfo{person}{Errui Ding}, {and}
  \bibinfo{person}{Xinghao Ding}.} \bibinfo{year}{2019}\natexlab{}.
\newblock \showarticletitle{Look More Than Once: An Accurate Detector for Text
  of Arbitrary Shapes}. In \bibinfo{booktitle}{\emph{{IEEE} Conference on
  Computer Vision and Pattern Recognition, {CVPR}}}. \bibinfo{publisher}{{IEEE}
  Computer Society}, \bibinfo{pages}{10552--10561}.
\newblock


\bibitem[\protect\citeauthoryear{Zhang, Zhang, Shen, Yao, Liu, and Bai}{Zhang
  et~al\mbox{.}}{2016}]%
        {multi-orient}
\bibfield{author}{\bibinfo{person}{Zheng Zhang}, \bibinfo{person}{Chengquan
  Zhang}, \bibinfo{person}{Wei Shen}, \bibinfo{person}{Cong Yao},
  \bibinfo{person}{Wenyu Liu}, {and} \bibinfo{person}{Xiang Bai}.}
  \bibinfo{year}{2016}\natexlab{}.
\newblock \showarticletitle{Multi-oriented Text Detection with Fully
  Convolutional Networks}. In \bibinfo{booktitle}{\emph{{IEEE} Conference on
  Computer Vision and Pattern Recognition, {CVPR}}}. \bibinfo{publisher}{{IEEE}
  Computer Society}, \bibinfo{pages}{4159--4167}.
\newblock


\end{thebibliography}

\end{document}